\documentclass{article}

\PassOptionsToPackage{numbers,compress}{natbib}

 \usepackage[preprint]{neurips_2026}

\usepackage[utf8]{inputenc} 
\usepackage[T1]{fontenc}    
\usepackage{hyperref}       
\usepackage{url}            
\usepackage{booktabs}       
\usepackage{amsfonts}       
\usepackage{nicefrac}       
\usepackage{microtype}      
\usepackage{xcolor}         
\usepackage{amssymb}
\usepackage{graphicx}
\usepackage{float}
\usepackage{subcaption}
\usepackage{booktabs} 
\usepackage{multirow}
\usepackage{array}
\newcolumntype{P}[1]{>{\raggedright\arraybackslash}p{#1}}
\usepackage{graphicx}
\usepackage{pifont}
\usepackage{caption}
\usepackage{wrapfig}
\usepackage{enumitem}
\usepackage[table]{xcolor}
\usepackage{amsmath}
\usepackage[ruled,vlined]{algorithm2e}
\usepackage{algorithmic}
\usepackage{makecell}
\usepackage{wrapfig}
\usepackage[ruled,vlined]{algorithm2e}
\usepackage{placeins}
\usepackage{bm}

\title{MindAlign: Bridging EEG, Vision, and Language for Zero-Shot Visual Decoding}

\author{
\textbf{Zexuan Chen\textsuperscript{1$\dagger$}} \quad
\textbf{Sichao Liu\textsuperscript{1,2,3$\dagger$,$\ddagger$}} \quad
\textbf{Runhao Lu\textsuperscript{2,4}} \quad
\textbf{Huichao Qi\textsuperscript{5}} \quad
\textbf{Alexandra Woolgar\textsuperscript{2}} \quad  \\
\textbf{Xi Vincent Wang\textsuperscript{1}}
\textbf{Lihui Wang\textsuperscript{1}} \\
\textsuperscript{1}KTH, SWeden \quad
\textsuperscript{2}University of Cambridge, UK \quad
\textsuperscript{3}EPFL, Switzerland \quad \\
\textsuperscript{4}McGill University, Canada \quad
\textsuperscript{5}Karolinska Institutet, Sweden
}

\begin{document}

\maketitle
\begingroup
\renewcommand\thefootnote{}
\footnotetext{$\dagger$: equal contributors;$\ddagger$ Corresponding author: \url{sicliu@kth.se}}
\endgroup

\begin{abstract}
Visual decoding from brain signals is a key challenge at the intersection of computer vision and neuroscience, requiring methods that bridge neural representations and computational models of vision. A field-wide goal is to achieve accurate, generalizable decoding from non-invasive, temporally resolved signals, including electroencephalography (EEG). A major obstacle towards this goal is the low signal-to-noise ratio of EEG and the substantial inter-subject variability, which render direct end-to-end EEG–image supervision weak and unstable. To address this, we introduce a tri-modal contrastive framework for EEG-based visual decoding that aligns EEG, visual, and textual representations within a unified latent space. Our approach follows a two-stage design. First, we pre-train an EEG encoder via masked reconstruction on unlabeled trials, learning spatio-temporal regularities that transfer robustly to downstream tasks. Second, we jointly align EEG, image, and LLM-generated textual descriptions through contrastive learning, where text supervision acts as a semantic regularizer that injects linguistic structure into the shared space without overwhelming the primary EEG–image signal. The encoder integrates subject-specific adaptation, graph-attention over channels, and temporal-spatial convolutional embeddings. On the Things-EEG2 200-way zero-shot benchmark, our framework achieves 54.1\% Top-1 and 83.4\% Top-5 accuracy, substantially exceeding the strongest prior baseline (32.4\% / 64.0\%), with paired Wilcoxon tests confirming significance (p < 0.01) over all in-subject baselines. We validate generalization on Things-MEG. Analysis reveals that compact embedding geometries (CN-CLIP) outperform much larger backbones, and that decoding aligns with established neurophysiology of visual processing. This work is a critical step towards robust, semantically-grounded visual decoding from non-invasive temporal neural signals. The source code is publicly available in an \href{https://github.com/anon-eeg/eeg_image_decoding}{\url{https://github.com/anon-eeg/eeg_image_decoding}}.


\end{abstract}


\section{INTRODUCTION}
Neuroscience has historically advanced through highly specialized studies of cognitive functions, resulting in a fragmented landscape of task-specific decoders tailored to individual experimental paradigms~\cite{mathis2024decoding,yamins2016using}. Developing robust brain-computer interfaces (BCIs) requires accurate, generalizable models of human visual processing from non-invasive neural signals. A major step in this direction has been the development of high-fidelity visual decoders of brain activities~\cite{kamitani2005Decoding,kay2008identifying,miyawaki2008visual} with recent advances further accelerated by contrastive multimodal learning and high-quality visual neuroimaging datasets. Visual decoding from electroencephalography (EEG) provides a particularly demanding testbed, as models must extract semantic information from signals that are noisy, temporally entangled, and spatially diffuse. The growing evidence that deep network latent hierarchies converge with the representational geometry of the human brain~\cite{yamins2014performance,kriegeskorte2015deep,caucheteux2022brains} has driven a wave of methods that align AI features trained with neural activities. The problem typically decomposes into two subproblems: (1) \emph{mapping high-dimensional, low-SNR neural activity to a compact visual-semantic representation}, and (2) \emph{aligning that representation with pretrained vision-language embedding spaces for recognition or retrieval}. Large-scale vision-language models~\cite{radford2021learning,liu2024improved,wang2024qwen2} have largely addressed (2), while large-scale EEG datasets~\cite{gifford2022large,grootswagers2022human} have driven recent progress on (1)~\cite{song2023decoding,song2025recognizing,li2024visual,wu2025bridging,li2025neural}.

Despite this progress, \textbf{a critical barrier still limits EEG-based decoding accuracy}. The low signal-to-noise ratio (SNR) of EEG and substantial inter-subject variability~\cite{darvas2004mapping,amin2017classification,puce2017review} make end-to-end EEG—image supervision weak and unstable, often yielding representations that fail to capture the richness of natural visual content and forcing per-subject models that cannot aggregate patterns across populations~\cite{haxby2011common}. Beyond signal noise, previous approaches also rely on pairwise EEG—image contrast or indirect semantic-space regression~\cite{song2025recognizing,du2023decoding}, leaving structured linguistic semantics — a complementary source of supervision — remains underexplored. More recent work — large-scale masked pre-training~\cite{ouahidi2025reve}, hierarchical channel-topology modeling~\cite{yang2025thd}, and continual subject adaptation~\cite{zhou2025spiced} — addresses these issues in isolation. Still, none combine self-supervision, structured channel modeling, and language-grounded supervision within a single framework. We address this gap by viewing EEG-based visual decoding as a cross-modal alignment problem in which a shared semantic space captures both visual appearance and linguistic meaning, and formulate decoding in two stages: (i) pre-train the EEG encoder with a masked reconstruction objective on unlabeled EEG, and (ii) transfer the pre-trained encoder and jointly align EEG, image, and LLM-generated text embeddings through contrastive learning. Image synthesis from brain activity is comparatively mature, we focus on decoding visual-semantic embeddings, evaluated through zero-shot image and text retrieval. \textbf{Our framework achieves 54.1\% Top-1 / 83.4\% Top-5 accuracy in the 200-way zero-shot setting on Things-EEG2}, versus \textbf{32.4\% / 64.0\%} for the strongest prior baseline.

\begin{figure}[!t]
    \centering
    \includegraphics[width=1\linewidth]{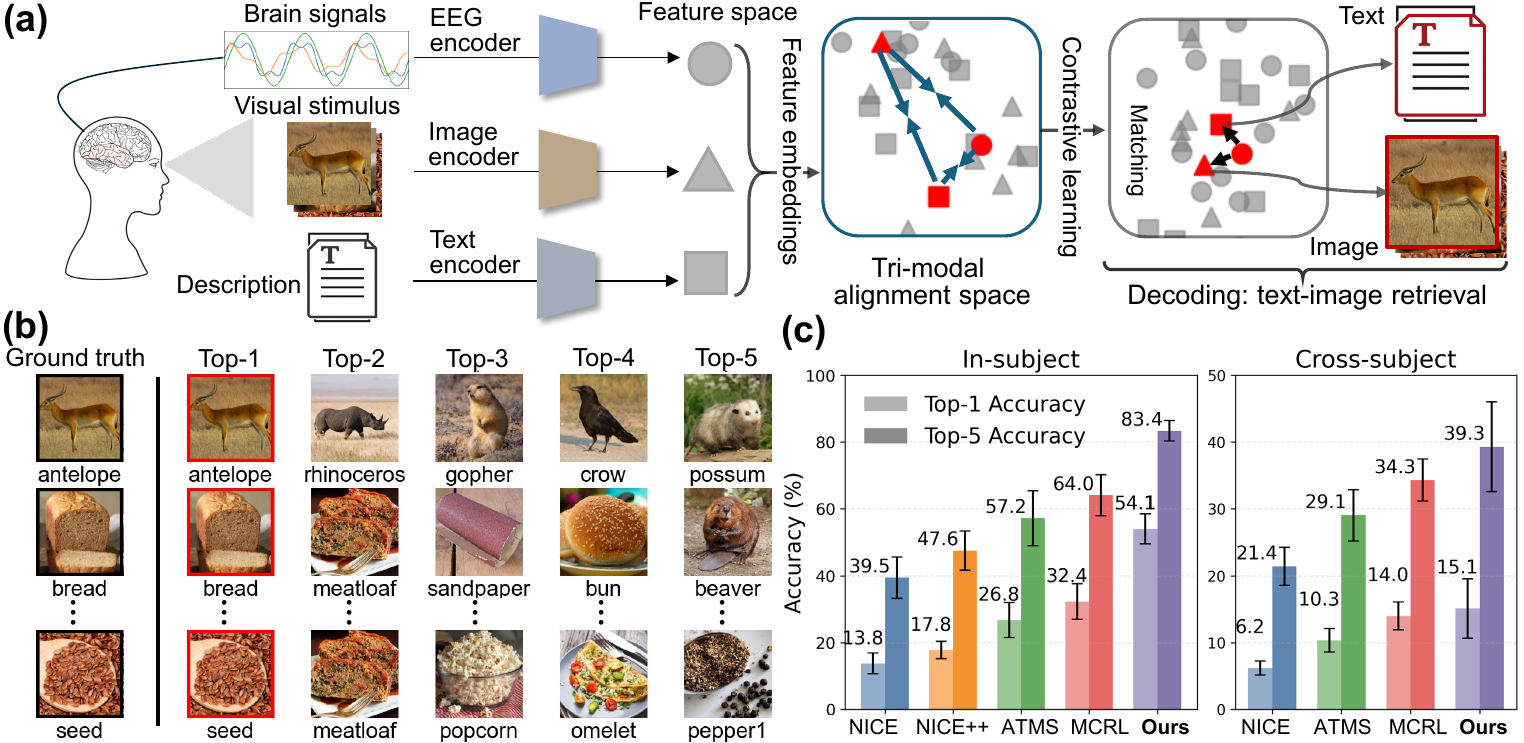}
    \caption{\textbf{Overview of the framework and decoding performance on Things-EEG2.} \textbf{(a)} Tri-modal contrastive alignment. EEG signals, visual stimuli, and LLM-generated descriptions are encoded into a shared feature space, where corresponding triplets are aligned through contrastive learning and mismatched samples are separated. At inference, an EEG embedding retrieves the most similar image and text candidates in this space. \textbf{(b)} Qualitative top-5 retrievals on the Things-EEG2 test set, ranked by EEG–image similarity; correct matches are highlighted. \textbf{(c)} Quantitative comparison against state-of-the-art (SoTA) baselines (NICE~\cite{song2023decoding}, NICE++~\cite{song2025recognizing}, ATMS~\cite{li2024visual}, MCRL~\cite{li2025neural}) under the in-subject (left) and cross-subject leave-one-subject-out (right) protocols, reporting Top-1 and Top-5 retrieval accuracy averaged over 10 participants. Our method leads in both settings, with the largest margin in-subject.}
    \label{fig:outline}
    \vspace{-1.8em} 
\end{figure}

Figure~\ref{fig:outline} summarizes the framework and headline results. Our main contributions are as follows.
\begin{itemize}[itemsep=0pt, topsep=0pt, leftmargin=*]
    \item A \textbf{\emph{tri-modal EEG–image–text alignment framework}} aligning EEG representations with image features and LLM-generated text in a shared embedding space, where textual semantics provide complementary supervision and improve discriminability over pairwise EEG–image alignment.
    \item A \textbf{\emph{high-performance EEG encoder}} that integrates a subject-specific adaptation layer, graph-attention-based channel modeling, and temporal-spatial convolutional patch embeddings to capture inter-channel and temporal dependencies.
    \item An \textbf{\emph{Masked Autoencoder (MAE)-based pre-training strategy}} that initializes the EEG encoder with masked reconstruction and partially transfers weights to the alignment stage, yielding consistent gains. We further observe that the \emph{geometry} of the visual target space may play an important role, with more compact embedding spaces outperforming larger backbones for EEG-to-image retrieval.
\end{itemize}

\section{RELATED WORKS}

\textbf{Contrastive Multimodal Learning for Visual Neural Decoding.}
Visual neural signal analysis follows two complementary paradigms~\cite{naselaris2011encoding}: \emph{encoding} models predict neural activity from stimuli~\cite{huth2016natural,schrimpf2018brain,caucheteux2022brains}, while \emph{decoding} models reconstruct or identify stimuli from neural activity~\cite{kamitani2005Decoding,song2023decoding,li2024visual,li2025neural,du2023decoding,spampinato2017deep}. Both have benefited from contrastive objectives~\cite{radford2021learning} that align neural activity with pretrained vision-language embeddings, motivated by the convergence between deep network hierarchies and the primate visual system~\cite{yamins2014performance,kriegeskorte2015deep}. CLIP~\cite{radford2021learning} has since been applied to fMRI~\cite{benchetrit2023brain,scotti2024mindeye2,wang2024mindbridge} and EEG~\cite{song2023decoding,li2024visual,song2025recognizing,wu2025bridging,li2025neural}. For EEG specifically, prior work has used coarse text labels as auxiliary supervision~\cite{spampinato2017deep} or indirect semantic-space regression~\cite{du2023decoding}. With the emergence of multimodal LLMs such as LLaVA-1.5~\cite{liu2024improved,liu2023visual} and Qwen2-VL~\cite{wang2024qwen2}, rich textual descriptions are now readily available. \textbf{We extend this line from pairwise EEG–image alignment to joint tri-modal EEG–image–text alignment, in which LLM-generated descriptions serve as an explicit third modality rather than label proxies}. A concurrent line of work targets deployment efficiency: ENIGMA~\cite{kneeland2025enigma} pairs subject-specific layers with a unified backbone for THINGS-EEG2 reconstruction. ENIGMA optimizes the parameter count under pairwise supervision, but \textbf{we enrich the supervisory signal itself with LLM-generated text}.

\textbf{Latent Space-based EEG Encoding.}
Discriminative EEG representations require jointly modeling sensor-level spatial dependencies and millisecond-scale temporal dynamics. Prior work has explored these axes largely in isolation: convolutional networks for spatially structured features~\cite{schirrmeister2017deep}, LSTMs for sequential dynamics~\cite{wang2018lstm,spampinato2017deep}, graph-based methods for inter-channel connectivity~\cite{zhong2020eeg,demir2021eeg,velivckovic2017graph,brody2021attentive}, attention-based parameterizations~\cite{adeli2023predicting,beliy2024wisdom} including temporal-spatial convolution (TSConv)~\cite{song2023decoding} and iTransformer variants treating each channel as a token~\cite{song2023decoding,li2024visual}, and subject-aware strategies for inter-individual variability~\cite{song2023decoding,song2025recognizing,benchetrit2023brain,guo2025neuro}. \textbf{We instead integrate these directions into a unified encoder that combines subject-specific adaptation, graph-attention-based channel modeling, Transformer-based global interactions, and temporal-spatial convolutional embeddings}. A recent work corroborates two of these design choices: THD-BAR~\cite{yang2025thd} imposes a multi-scale spatial hierarchy on channels to overcome the limits of purely time-centered modeling, and SPICED~\cite{zhou2025spiced} addresses inter-subject variability through bio-inspired continual adaptation.

\textbf{Self-Supervised Pre-training via Masked Reconstruction.}
Self-supervised learning extracts transferable representations from unlabeled data~\cite{he2022masked,devlin2019bert,hospedales2021meta} and underpins foundation models in neuroscience~\cite{schneider2023learnable,wang2025foundation,binz2025foundation,d2025tribe}. MAE~\cite{he2022masked} and BERT~\cite{devlin2019bert} establish masked reconstruction as a cross-modal paradigm with modality-tailored masking ratios. Early EEG adaptations~\cite{chien2022maeeg,bai2024dreamdiffusion,wang2025eegmamba} apply random temporal masking for classification under limited supervision; REVE~\cite{ouahidi2025reve} recently scaled MAE pre-training to 60{,}000 hours and 25{,}000 subjects, establishing it as the dominant EEG self-supervision paradigm. NeurIPT~\cite{fang2025neuript} further shows that EEG-specific masking outperforms vision/language defaults — consistent with our finding (Sec.~\ref{sec:ablation}) that the optimal ratio for EEG sits between the vision and language extremes. These approaches treat pre-training and downstream supervision as loosely coupled. In contrast, \textbf{our encoder is explicitly designed for partial weight transfer — particularly of its subject-specific layer — into the alignment stage, providing more robust initialization and consistent downstream gains}.

\section{METHODS}
\label{sec:methods}
\textbf{Problem Definition.}\label{sec:motivation_problem}
The low SNR of EEG and substantial inter-subject variability pose a major obstacle to accurate, generalizable visual decoding from neural signals. Rather than learning a fixed mapping from EEG to visual embeddings under such weak supervision, we formulate \textbf{EEG-based visual decoding as a cross-modal alignment problem that grounds noisy neural signals in a shared visual–semantic space, without requiring explicit category-level supervision at test time}.

\begin{wrapfigure}{R}{0.58\linewidth}
    \centering
    \includegraphics[width=\linewidth]{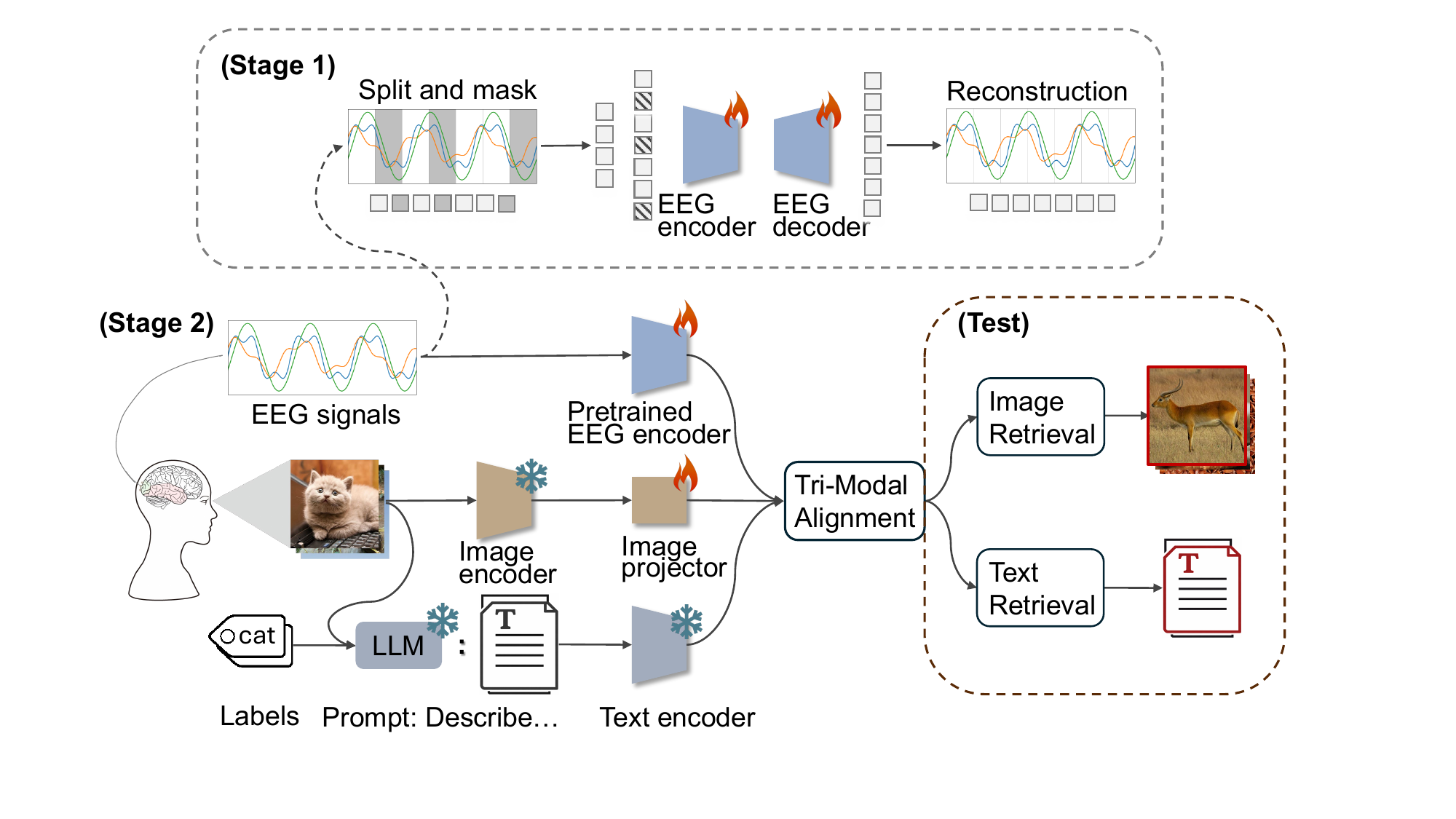}
    \caption{\textbf{Framework overview.} \textbf{Stage 1 (pre-training):} EEG signals are split and partially masked with noise, and reconstructed by a lightweight decoder from the encoder's latents, driving the encoder to learn intrinsic neural dynamics. \textbf{Stage 2 (tri-modal alignment):} the pre-trained EEG encoder is jointly trained with frozen image and text encoders, where text descriptions are generated by an LLM from visual content and category labels. Two contrastive losses — EEG–image and image–text — align all three modalities in a shared space. At test time, EEG embeddings retrieve images and text via cross-modal similarity.}
    \vspace{-1.8em}
    \label{fig:overview}
\end{wrapfigure}

Let an image $I$ be encoded as $\mathbf{F}_{\text{img}} = \phi_{\text{img}}(I) \in \mathbb{R}^{1 \times d}$ by a frozen pretrained image encoder $\phi_{\text{img}}$ (e.g., CLIP), where $d$ is the shared embedding dimension. Let $\mathbf{F}_{\text{text}} = \phi_{\text{text}}(\mathcal{D}(I, c)) \in \mathbb{R}^{1 \times d}$ denote the embedding of an LLM-generated description $\mathcal{D}(I, c)$ conditioned on $I$ and its category label $c$, $\phi_{\text{text}}$ is the frozen text encoder paired with $\phi_{\text{img}}$ from the same pretrained CLIP model. For each stimulus $I$, the EEG response of subject $s$ is $\mathbf{X} \in \mathbb{R}^{C \times T}$ ($C$ channels, $T$ time samples). At test time, given only the EEG response $\mathbf{X}'$ from subject $s$ to an unseen stimulus, \textbf{our goal is to infer its visual–semantic embedding $\mathbf{F}_{\text{img}}^{\text{novel}}$ via cross-modal similarity in the shared space}.

Realizing this formulation requires an EEG representation that is robust to noise and inter-subject variability, and aligned with both visual and linguistic semantics. We propose a two-stage tri-modal framework (Fig.~\ref{fig:overview}). \textbf{Stage 1: Pre-training.} The EEG encoder is pre-trained by masked reconstruction — spatio-temporal patches are replaced with noise, and a lightweight decoder reconstructs the original signal from the encoder's latents. The decoder is discarded and the encoder weights are transferred to Stage 2. \textbf{Stage 2: Tri-modal alignment.} The pre-trained encoder forms the EEG branch. The image branch applies a trainable projection on a frozen image encoder. The text branch prompts an LLM with \emph{``Describe only what is directly visible in the image of $\langle$label$\rangle$ in one short sentence''}, encoded by a frozen text encoder. Two contrastive losses — EEG–image and image–text — are jointly optimized. The shared image representation serves as an intermediate reference that implicitly aligns EEG with text. At inference, all modules are frozen: EEG embeddings are matched to image candidates in the shared space (200-way zero-shot), with text retrieval as auxiliary. The framework can robustly transfer and generalize to MEG.

\vspace{-0.5em}
\subsection{EEG Encoder Design}\label{sec:EEG_encoder}
The EEG encoder maps a minibatch $\mathbf{X} \in \mathbb{R}^{B \times C \times T}$ ($B$: batch, $C$: channels, $T$: time samples) to a $d$-dimensional representation $\mathbf{F}_{\text{eeg}} \in \mathbb{R}^{B \times d}$ aligned with the visual–semantic space. It consists of five components applied sequentially: (i) a subject-specific adaptation layer, (ii) a Graph Attention Network (GAT) for local inter-channel coupling, (iii) a Transformer for global channel-level interactions, (iv) channel-wise attention with spatial-electrode priors, and (v) a temporal-spatial convolutional patch embedding. Each stage produces an intermediate tensor $\mathbf{X}_k$ ($k=1,\dots,5$) preserving the $(B, C, T)$ shape until the final patch embedding and projection.

\textbf{(i) Subject-specific adaptation.} To absorb inter-subject variability, a learnable transformation $\mathbf{W}_s$ is applied per subject $s$, producing $\mathbf{X}_1 = \mathbf{W}_s \mathbf{X} \in \mathbb{R}^{B \times C \times T}$.

\textbf{(ii) Graph Attention Network.} EEG channels are treated as nodes in a fully connected graph~\cite{velivckovic2017graph, brody2021attentive}. Letting $\mathbf{h}_i \in \mathbb{R}^{T}$ be the temporal sequence at channel $i$ (the $i$-th row of $\mathbf{X}_1$), $\mathcal{N}(i)$ be the set of neighbors of node $i$. The node updates and the normalized attention coefficient $\alpha_{ij}$ are defined as:
\begin{equation}
\mathbf{h}_i' = \sum_{j \in \mathcal{N}(i)} \alpha_{ij}\, \mathbf{W}_g \mathbf{h}_j,
\quad
\alpha_{ij} = \frac{\exp\!\bigl(\mathrm{LeakyReLU}(\mathbf{a}^{\top}[\mathbf{W}_g\mathbf{h}_i \,\Vert\, \mathbf{W}_g\mathbf{h}_j])\bigr)}
{\sum_{k \in \mathcal{N}(i)} \exp\!\bigl(\mathrm{LeakyReLU}(\mathbf{a}^{\top}[\mathbf{W}_g\mathbf{h}_i \,\Vert\, \mathbf{W}_g\mathbf{h}_k])\bigr)},
\end{equation}
where $\mathbf{W}_g$ is a learnable projection and $\mathbf{a}$ is a learnable attention vector. A residual connection yields $\mathbf{X}_2 = \mathrm{GAT}(\mathbf{X}_1) + \mathbf{X}_1$.

\textbf{(iii) Transformer over channel tokens.} While the GAT performs attention over a graph structure, and the Transformer captures \emph{global} dependencies via dense self-attention.
We treat $\mathbf{X}_2$ as a sequence of $C$ channel tokens, project them to a latent space $\mathbf{Z}_0 = \mathrm{Embedding}(\mathbf{X}_2)$, and apply self-attention across channels:
\begin{equation}
\mathrm{Attention}(\mathbf{Z}_0) = \mathrm{Softmax}\!\left(\frac{\mathbf{Q}\mathbf{K}^{\top}}{\sqrt{d_k}}\right)\mathbf{V},
\end{equation}
where $\mathbf{Q}$, $\mathbf{K}$, and $\mathbf{V}$ are linear projections of $\mathbf{Z}_0$, and $d_k$ is the key dimension for scaling.
A stack of Transformer layers with residual connections produce $\mathbf{X}_3$.

\textbf{(iv) Channel-wise attention with spatial priors.} A two-layer MLP applied to temporally pooled features (mean over $T$) produces channel-wise gating weights, giving the reweighted representation $\mathbf{X}_4 = \mathbf{X}_3 \odot \sigma\bigl(\mathrm{MLP_1}(\mathrm{Pool}_T(\mathbf{X}_3))\bigr) + \mathbf{X}_3$.
We further inject anatomical structure using standardized 3D electrode coordinates~\cite{seeck2017standardized}, augmented with radial distance and embedded via another MLP: $\mathbf{X}_5 = \mathbf{X}_4 + \mathrm{Proj}\bigl(\mathrm{MLP_2}([\mathbf{coords},\, \lVert\mathbf{coords}\rVert_2])\bigr)$

\textbf{(v) Temporal-spatial patch embedding and projection.} Following~\cite{song2023decoding}, $\mathbf{X}_5$ is normalized and encoded through a temporal-spatial convolutional patch embedding, then mapped to the shared $d$-dimensional space by a linear projection head, yielding $\mathbf{F}_{\text{eeg}} \in \mathbb{R}^{B \times d}$.

\vspace{-0.6em}
\subsection{Mask-Reconstruction Pre-training}\label{sec:mask_reconstruction}
\vspace{-0.6em}
EEG–image pairing alone provides weak supervision: the low SNR of EEG and the limited number of paired trials make the contrastive objective unstable. We mitigate this by pre-training the EEG encoder with a self-supervised masked-reconstruction objective inspired by the MAE~\cite{he2022masked}, which encourages the encoder to learn intrinsic spatio-temporal regularities of EEG signals.

\textbf{Patchification and masking.}
Given $\mathbf{X} \in \mathbb{R}^{B \times C \times T}$, we partition each sample along the time axis into $L = T/p$ non-overlapping patches of length $p$, yielding $\mathbf{X}_{\text{patch}} \in \mathbb{R}^{B \times L \times (Cp)}$. A subset of patches is then selected uniformly at random according to a masking ratio $r$ and replaced with Gaussian noise; the remaining patches are kept unchanged. Unlike vision MAE, which uses learned mask tokens, we found Gaussian noise based corruption leads to more stable training for low-SNR EEG.

\textbf{Encoder–decoder reconstruction.}
The corrupted sequence is fed into the EEG encoder (Sec.~\ref{sec:EEG_encoder}) to obtain latent representations, which are projected to dimension $W$, augmented with positional embeddings, and processed by a lightweight Transformer decoder with $D$ layers. The decoder output $\mathbf{Z}_D$ is mapped back to patch space by a linear head parameterized by $\mathbf{W}_{\text{pred}}$ and $\mathbf{b}_{\text{pred}}$:
\begin{equation}
\widehat{\mathbf{X}}_{\text{patch}} = \mathbf{W}_{\text{pred}}\,\mathrm{LayerNorm}(\mathbf{Z}_D) + \mathbf{b}_{\text{pred}} \in \mathbb{R}^{B \times L \times Cp}.
\end{equation}

\textbf{Reconstruction loss.}
Reconstruction is supervised by patch-level mean squared error, averaged across both the $Cp$ channel-time entries and the $L$ patches per sample:
\begin{equation}
\textit{Loss} = \frac{1}{BLCp}\sum_{b=1}^{B}\sum_{i=1}^{L}\sum_{j=1}^{Cp} \bigl(\widehat{\mathbf{X}}_{\text{patch}}[b,i,j] - \mathbf{X}_{\text{patch}}[b,i,j]\bigr)^{2}.
\end{equation}
Notably, the loss is computed over \emph{all} patches rather than only masked ones. We observe empirically that reconstructing the full sequence stabilizes training on noisy EEG and yields more consistent spatio-temporal representations than the masked-only variant.

\textbf{Weight transfer.}
After pre-training, the decoder is discarded. All encoder weights, \emph{including the subject-specific adaptation layer}, are transferred to Stage 2. In Sec.~\ref{sec:multimodal_alignment}, transferring the subject-specific layer accounts for the majority of the gain.

\vspace{-0.8em}
\subsection{Multimodal Alignment}\label{sec:multimodal_alignment}
In Stage 2, we jointly align EEG, image, and text representations within the shared embedding space via two contrastive losses: 1) an EEG–image term that supplies the primary supervisory signal, and 2) an image–text term that injects linguistic structure into the shared space. As both EEG and text embeddings are pulled toward the same image representation, an EEG–text alignment emerges \emph{implicitly} without a third contrastive term.

\textbf{Cross-modal similarities.}
Following Algorithm~\ref{alg:tri_modal}, we $\ell_2$-normalize all three embeddings and compute EEG–image and image–text cosine similarity matrices $\mathbf{S}_{\text{EI}}$ and $\mathbf{S}_{\text{IT}}$, which are scaled by a learnable temperature $\tau$:
\begin{equation}
\mathbf{S}_{\text{EI}} = \tau\, {\mathbf{F}}_{\text{eeg}} \, {\mathbf{F}}_{\text{img}}^{\top}, \qquad
\mathbf{S}_{\text{IT}} = \tau\, {\mathbf{F}}_{\text{img}} \, {\mathbf{F}}_{\text{text}}^{\top}.
\end{equation}
\textbf{Symmetric InfoNCE objective.}
Define $\mathbf{y} = [1, 2, \dots, B]$, $y_i$ denotes the index of the corresponding positive sample for the $i$-th element in the batch. $\mathcal{L}_{\text{EI}}$ (EEG–image objective) is formulated as a symmetric InfoNCE loss, and $\mathcal{L}_{\text{IT}}$ (image–text objective) is defined analogously using $\mathbf{S}_{\text{IT}}$.
\begin{equation}
\mathcal{L}_{\text{EI}} = \tfrac{1}{2}\bigl(\mathcal{L}_{\text{CE}}(\mathbf{S}_{\text{EI}}, \mathbf{y}) + \mathcal{L}_{\text{CE}}(\mathbf{S}_{\text{EI}}^{\top}, \mathbf{y})\bigr), \quad
\mathcal{L}_{\text{CE}}(\mathbf{S}_{\text{EI}}, \mathbf{y}) = -\frac{1}{B}\sum_{i=1}^{B}
\log\frac{\exp(\mathbf{S}_{\text{EI}}[i, y_i])}
{\sum_{j=1}^{B}\exp(\mathbf{S}_{\text{EI}}[i, j])}.
\end{equation}
\textbf{Total objective.}
The final loss $\mathcal{L}_{\text{total}}$ is a convex combination weighted by $\alpha \in [0, 1]$:
\begin{equation}
\mathcal{L}_{\text{total}} = (1 - \alpha)\,\mathcal{L}_{\text{EI}} + \alpha\,\mathcal{L}_{\text{IT}}.
\label{eq:total_loss}
\end{equation}
A small $\alpha$ injects linguistic structure into the shared space without affecting EEG–image alignment. We set $\alpha = 0.1$ based on validation. 
Although larger $\alpha$ improves text retrieval, we use a smaller value to align with the prevailing focus on image retrieval.
In this setting, image–text supervision acts as a mild regularizer, enriching the embedding space with semantic structure.


\section{EXPERIMENTS AND RESULTS}
\subsection{Experimental Setup}
\label{sec:exp}
\vspace{-0.5em}

\begin{wrapfigure}{r}{0.5\textwidth}
\vspace{-2em}
\begin{algorithm}[H]
\footnotesize
\caption{Tri-modal Alignment Training}
\label{alg:tri_modal}
\KwIn{EEG signals, pre-extracted image features, and pre-extracted text features}
\KwOut{Trained EEG encoder and image projection head}

$\mathrm{EEG} \leftarrow$ tensor of shape $(B, C, T)$\;
$\mathrm{Image} \leftarrow$ pre-extracted features of shape $(B, d)$\;
$\mathrm{Text} \leftarrow$ pre-extracted features of shape $(B, d)$\;

\For{each batch}{
    ${\mathbf{F}}_{\text{eeg}} \gets \mathrm{Normalize}(\mathrm{EEG\_Enc}(\mathrm{EEG}))$\;
    ${\mathbf{F}}_{\text{img}} \gets \mathrm{Normalize}(\mathrm{Proj\_Img}(\mathrm{Image}))$\;
    ${\mathbf{F}}_{\text{text}} \gets \mathrm{Normalize}(\mathrm{Text})$\;

    $\mathcal{L}_{\text{EI}} \gets 
    \mathrm{Contrastive\_Loss}({\mathbf{F}}_{\text{eeg}}, 
    {\mathbf{F}}_{\text{img}})$\;

    $\mathcal{L}_{\text{IT}} \gets 
    \mathrm{Contrastive\_Loss}({\mathbf{F}}_{\text{img}}, 
    {\mathbf{F}}_{\text{text}})$\;

    $\mathcal{L}_{\text{total}} \gets 
    (1-\alpha)\mathcal{L}_{\text{EI}} + 
    \alpha \mathcal{L}_{\text{IT}}$\;

    Back-propagate $\mathcal{L}_{\text{total}}$ and update parameters\;
}
\end{algorithm}
\vspace{-2em}
\end{wrapfigure}
\textbf{Datasets.} We use two THINGS-based benchmarks. 
\textit{Things-EEG2}~\cite{gifford2022large}: 63-channel EEG from 10 
participants under a rapid serial visual presentation paradigm (200 ms stimulus onset asynchrony), with 1{,}654 training concepts (10 images $\times$ 4 repetitions) and 200 disjoint test concepts 
(1 image $\times$ 80 repetitions), defining a 200-way zero-shot retrieval 
task. \textit{Things-MEG}~\cite{hebart2023thingsdata}: 271-channel MEG from 
4 participants over 1{,}854 concepts, used for cross-modality validation (see specifications in Table~\ref{tab:datasets}, Appendix~\ref{app:dataset}).

\textbf{Evaluation.} For each test trial, we rank the 200 candidate image 
embeddings by cosine similarity and report Top-1 and Top-5 accuracy 
(chance: 0.5\% / 2.5\%). The \textit{in-subject} protocol trains and tests 
on the same participant; the \textit{cross-subject} protocol uses 
leave-one-subject-out (LOSO), where a single shared subject layer is trained on data aggregated from remaining subjects and tested on the held-out subject, serving as a shared adaptation module rather than a per-subject parameterization.

\textbf{Baselines.} We compare against NICE~\cite{song2023decoding}, 
NICE++~\cite{song2025recognizing}, ATMS~\cite{li2024visual}, 
MCRL~\cite{li2025neural}, UBP~\cite{wu2025bridging}. All baselines share the 
EEG preprocessing pipeline and 200-way zero-shot split for comparability.

\vspace{-0.5em}
\subsection{Overall Performance}
\vspace{-0.5em}

\begin{wrapfigure}{T}{0.71\linewidth}
    \centering
    \vspace{-1.3em}
    \includegraphics[width=\linewidth]{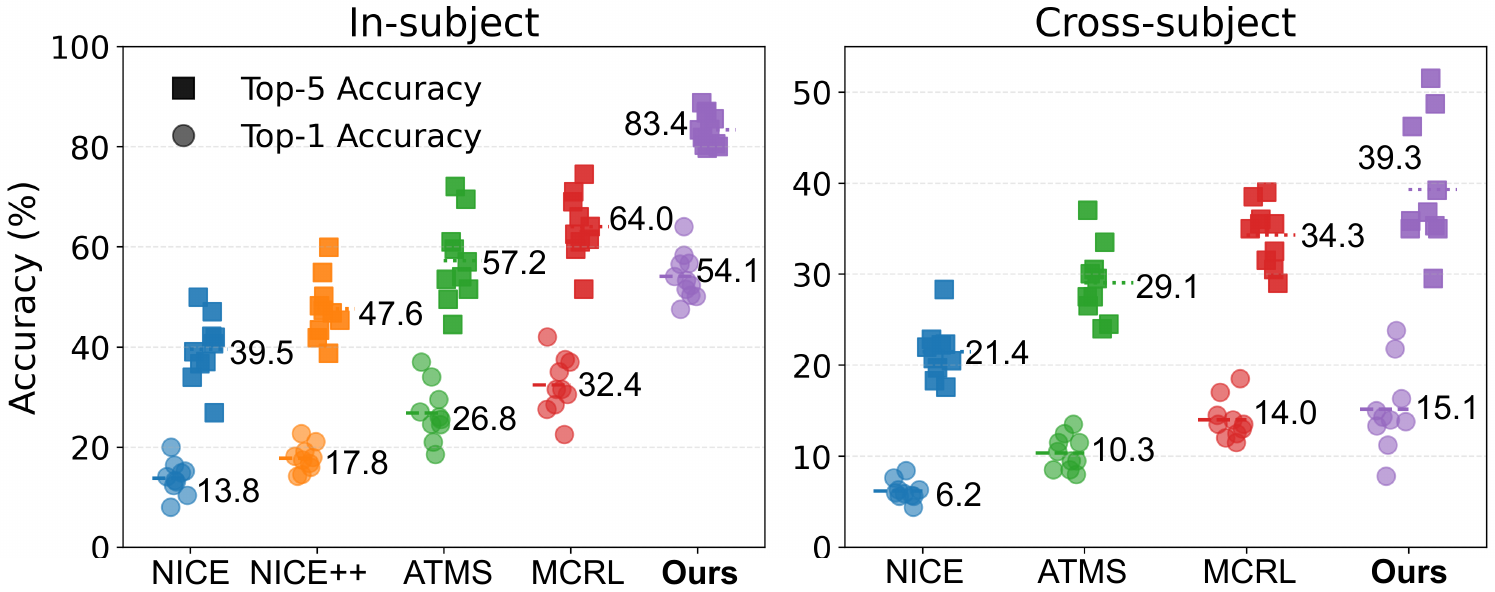}
    \vspace{-1.2em}
    \caption{\textbf{EEG decoding performance on the Things-EEG2 dataset}.
    \textbf{Left:} \textbf{In-subject} comparison across five methods;
    \textbf{Right:} \textbf{Cross-subject} (leave-one-subject-out) comparison
    across four methods. (see details in Tables~\ref{tab:accuracy_in_subject} \&~\ref{tab:accuracy_cross_subject} in Appendix~\ref{app:additional_results}, respectively) }
    \vspace{-1.8em}  
    \label{fig:overall}
\end{wrapfigure}

We evaluate our framework for EEG-to-image recognition under two protocols. In the \textit{in-subject} setting, the model is trained and tested on data from the same participant. In the \textit{cross-subject} setting, generalization is assessed using a LOSO protocol across all 10 subjects, where the model is trained on nine subjects and tested on the held-out one.
 
The results on Things-EEG2 are summarized in Fig.~\ref{fig:overall}. \textbf{In the in-subject setting, our model achieves a mean Top-1 accuracy of 54.1\% and a Top-5 accuracy of 83.4\%}, substantially outperforming recent SoTA methods, including \textbf{NICE} (Top-1: 12-20\%, Top-5: 27-50\%)~\cite{song2023decoding}, \textbf{NICE++} (Top-1: 14--23\%, Top-5: 39--60\%)~\cite{song2025recognizing}, \textbf{ATMS} (Top-1: 18-37\%, Top-5: 44-72\%)~\cite{li2024visual}, and \textbf{MCRL} (Top-1: 22-42\%, Top-5: 51-74\%)~\cite{li2025neural}. \textbf{Tests over the 10 per-subject Top-1 \& Top-5} scores show \textbf{statistically significant improvements over all baselines} (\(p < 0.01\) vs.\ NICE, ATMS, and MCRL). These results demonstrate that our framework captures richer visual-semantic representations from EEG than approaches relying solely on EEG-image pairing.

In the cross-subject setting, the performance of all methods declines because of substantial inter-subject variability. Nevertheless, \textbf{our model maintains strong decoding capability, achieving Top-1 accuracies in the range of $7.8\%$-$23.8\%$ and Top-5 accuracies of $29.5\%$-$51.5\%$}, while consistently outperforming competing approaches across subjects ($p < 0.01$ vs.\ each baseline on Top-1/Top-5, except for MCRL on Top-1 where $p=0.084$). These results indicate that our method captures subject-specific neural signatures and also transfers to unseen participants, 
demonstrating strong within-subject modeling and cross-subject generalization capabilities.

\begin{wraptable}{r}{0.62\textwidth}
\vspace{-1em}
\centering
\scriptsize
\setlength{\tabcolsep}{1.5pt}
\caption{\textbf{Top-1 and Top-5 image retrieval accuracy} (\%) on the \textbf{{\emph{MEG dataset}}} under in-subject and cross-subject settings.}
\resizebox{0.62\textwidth}{!}{
\begin{tabular}{lcccccccccc}
\toprule
\multirow{2}{*}{Method}
& \multicolumn{2}{c}{Subject1}
& \multicolumn{2}{c}{Subject2}
& \multicolumn{2}{c}{Subject3}
& \multicolumn{2}{c}{Subject4}
& \multicolumn{2}{c}{Average} \\

\cmidrule(lr){2-3}
\cmidrule(lr){4-5}
\cmidrule(lr){6-7}
\cmidrule(lr){8-9}
\cmidrule(lr){10-11}

& Top-1 & Top-5
& Top-1 & Top-5
& Top-1 & Top-5
& Top-1 & Top-5
& Top-1 & Top-5 \\
\midrule

NICE~\cite{song2023decoding}
& 6.9 & 20.5 & 15.3 & 37.1 & 12.3 & 35.0 & 5.8 & 21.1 & 10.1 & 28.4 \\

NICE++~\cite{song2025recognizing}
& 8.1 & 22.9 & 17.3 & 42.7 & 14.2 & 40.2 & 7.5 & 23.9 & 11.8 & 32.4 \\

\textbf{Ours}
& \textbf{9.2} & \textbf{31.7}
& \textbf{45.2} & \textbf{80.8}
& \textbf{32.2} & \textbf{65.8}
& \textbf{14.7} & \textbf{37.2}
& \textbf{25.3} & \textbf{53.9} \\

\midrule
\multicolumn{11}{c}{\textbf{Cross-Subject}} \\
\midrule

UBP~\cite{wu2025bridging}
& 2.0 & 5.7  & 1.5 & 17.2 & \textbf{2.7} & 10.5 & \textbf{2.5} & 8.0  & 2.2 & 10.4 \\

\textbf{Ours}
& \textbf{2.7} & \textbf{8.7}
& \textbf{4.7} & \textbf{18.5}
& 2.5 & \textbf{11.7}
& 1.7 & \textbf{10.8}
& \textbf{2.9} & \textbf{12.4} \\

\bottomrule
\end{tabular}
}
\label{tab:meg_comparison}
\vspace{-1em}
\end{wraptable}

We evaluate on Things-MEG (Table~\ref{tab:meg_comparison}): in the 
in-subject setting, our model substantially outperforms NICE and NICE++ 
(Top-1: $25.3\%$ vs.\ $11.8\%$; Top-5: $53.9\%$ vs.\ $32.4\%$); in the 
more challenging cross-subject setting, it achieves the best average 
performance ($2.9\%$ Top-1 / $12.4\%$ Top-5), surpassing 
UBP~\cite{wu2025bridging}. The results confirm that the framework transfers from EEG to MEG without architectural redesign, with only modality-dependent parameter adjustments (e.g., channel number).
We also evaluate our model on the complementary text-retrieval task while 
varying the alignment weight $\alpha$, which balances the EEG--image and 
image--text contrastive objectives. As detailed in 
Table~\ref{tab:text_retrieval_alpha} (Appendix~\ref{app:additional_results}), 
average Top-1 accuracy rises from $8.4\%$ at $\alpha{=}0.1$ to $11.8\%$ 
at $\alpha{=}0.7$, and Top-5 from $26.4\%$ to $32.2\%$, reflecting a 
complementary trade-off between the two alignment pathways.

\vspace{-0.5em}
\subsection{Ablation study}
\label{sec:ablation}
\vspace{-0.5em}

\begin{wraptable}{r}{0.65\textwidth}
\vspace{-1em}
\centering
\small
\setlength{\tabcolsep}{2pt}
\renewcommand{\arraystretch}{1.12}
\caption{\textbf{Pre-training ablation results}. \textbf{Top-1 accuracy averaged across 
10 subjects} (mean $\pm$ std.)}
\label{tab:mae_ablation}
\resizebox{0.65\textwidth}{!}{
\begin{tabular}{lcccc}
\toprule
Decoder (W, D) & $r=0.15$ & $r=0.3$ & $r=0.5$ & $r=0.75$ \\
\midrule
(256,2) & 52.58 $\pm$ 8.06 & \textbf{53.93 $\pm$ 6.76} & 52.40 $\pm$ 6.59 & 52.85 $\pm$ 7.51 \\
(512,4) & 52.95 $\pm$ 7.13 & 53.67 $\pm$ 7.12 & 53.83 $\pm$ 6.69 & 51.45 $\pm$ 8.39 \\
(512,8) & 52.60 $\pm$ 7.16 & 53.33 $\pm$ 7.51 & 52.97 $\pm$ 7.19 & 52.42 $\pm$ 7.01 \\
\bottomrule
\end{tabular}
}
\vspace{-1.0em}
\end{wraptable}
\textbf{MAE-based Pre-training Configuration.} Following the MAE design~\cite{he2022masked}, a lightweight 
Transformer decoder is attached during pre-training and discarded 
afterward, allowing the encoder and decoder to be sized independently. We sweep three hyperparameters: decoder width $W$, depth $D$, and 
masking ratio $r$ (Table~\ref{tab:mae_ablation}). The best mean Top-1 
accuracy ($53.93\%$) is obtained with $(W{=}256, D{=}2, r{=}0.3)$. 
While absolute differences across configurations fall within one standard 
deviation, $r{=}0.3$ is consistently among the top results across all 
decoder sizes, and larger decoders yield no clear gain. We 
adopt the smallest decoder with $r{=}0.3$ for both efficiency and 
robustness.

\begin{wraptable}{r}{0.65\textwidth}
\vspace{-1em}
\centering
\small
\setlength{\tabcolsep}{2pt}
\renewcommand{\arraystretch}{1.12}
\caption{\textbf{Component ablation}. Each column reports performance when the 
indicated module is removed; \emph{Spatial-spectral} jointly denotes the 
channel-wise attention and spatial channel embeddings.}
\label{tab:ablation}
\resizebox{0.65\textwidth}{!}{
\begin{tabular}{lccccc}
\toprule
Module & Spatial-spectral & Subject layer & Transformer & GAT & Full Model \\
\midrule
Top-1 & 53.0 $\pm$ 7.8 & 52.1 $\pm$ 6.8 & 53.8 $\pm$ 7.2 & 53.3 $\pm$ 7.3 & \textbf{54.1 $\pm$ 4.9} \\
Top-5 & 82.7 $\pm$ 5.4 & 82.3 $\pm$ 4.7 & 84.4 $\pm$ 4.0 & 83.6 $\pm$ 4.4 & 83.4 $\pm$ 3.4 \\
\bottomrule
\end{tabular}
}
\vspace{-1.0em}
\end{wraptable}
The optimal $30\%$ ratio lies between the $75\%$ used in vision 
MAE~\cite{he2022masked} and the $15\%$ used in BERT~\cite{devlin2019bert}, 
reflecting EEG's intermediate redundancy: strong spatial correlations 
across nearby electrodes and temporal continuity in neural activity, 
yet sensitivity to fine-grained stimulus-locked structure. Aggressive 
masking destroys informative patterns; conservative masking fails to 
elicit context modeling. A $30\%$ ratio balances both regimes, 
suggesting EEG masking is tailored to its signal properties 
rather than inherited from vision/language defaults.

\textbf{EEG Encoder.} We ablate the EEG encoder along two axes: (i) removing individual modules 
from the full architecture, and (ii) disabling pre-trained weight transfer 
for specific components. As shown in Table~\ref{tab:ablation}, removing the subject-specific layer 
produces the highest drop in Top-1 (-2.0\%), confirming that 
inter-subject variability is the dominant factor in EEG-based visual 
decoding. The remaining modules contribute smaller, overlapping gains; 
notably, the Full Model achieves the lowest variance (std $4.9$ vs.\ 
$6.8$--$7.8$ for ablated variants), indicating that the combined design 
yields more stable representations across subjects.

\begin{wraptable}{r}{0.55\textwidth}
\centering
\small
\setlength{\tabcolsep}{3pt}
\renewcommand{\arraystretch}{1.12}
\caption{\textbf{Effect of pre-training transfer strategies}.}
\label{tab:pretrain_ablation}
\resizebox{0.55\textwidth}{!}{
\begin{tabular}{lccc}
\toprule
Strategy & None & All except Subject Layer & All Components \\
\midrule
Top-1 (\%) & 52.18 $\pm$ 6.90 & 52.70 $\pm$ 6.87 & \textbf{54.05 $\pm$ 4.87} \\
Top-5 (\%) & 82.37 $\pm$ 4.39 & 82.50 $\pm$ 4.51 & \textbf{83.37 $\pm$ 3.39} \\
\bottomrule
\end{tabular}
}
\vspace{-1em}
\end{wraptable}

Table~\ref{tab:pretrain_ablation} compares three transfer strategies: 
no pre-training, transferring all weights except the subject-specific 
layer, and transferring all weights. Transferring \emph{all} weights yields the largest gain ($+1.9$ 
Top-1 over training from scratch). 
The subject-specific layer benefits most from pre-training, likely because masked reconstruction learns informative channel-level representations, providing a strong initialization for subject-specific adaptation.

\textbf{Image encoder.}
We adopt the same CLIP backbone for both image and text encoders and compare four CLIP models spanning two orders of magnitude in size
(see Table~\ref{tab:vision_encoder_overview}, 
Appendix~\ref{app:preprocessing}). CN-CLIP (RN50, 38M parameters) outperforms CLIP-ViT-G-14 (1.37B) by 16.7\% Top-1 
despite being 36$\times$ smaller. 
Results are shown in Tables~\ref{tab:model_comparison} and \ref{tab:image_encoder_subjects} (Appendix~\ref{app:additional_results}).
We hypothesize two contributing 
factors: the ResNet backbone's locality bias may better match the coarse, 
low-SNR structure of EEG, and CN-CLIP's smaller, more curated training 
corpus ($\sim$200M pairs) may yield a more compact embedding geometry 
better suited to contrastive alignment with limited EEG signal.

\begin{wraptable}{r}{0.55\textwidth}
\vspace{-1.2em}
\centering
\small
\setlength{\tabcolsep}{3pt}
\renewcommand{\arraystretch}{1.12}
\caption{\textbf{Image retrieval results across CLIP vision-language backbones}
(mean $\pm$ std over 10 subjects).}
\label{tab:model_comparison}
\resizebox{0.55\textwidth}{!}{
\begin{tabular}{lcccc}
\toprule
Metric & ViT-L-14~\cite{dosovitskiy2020image} & ViT-H-14~\cite{dosovitskiy2020image} & ViT-G-14~\cite{zhai2022scaling} & \textbf{CN-CLIP}~\cite{yang2022chinese} \\
\midrule
Top-1 (\%) & 39.6 $\pm$ 8.0 & 41.3 $\pm$ 7.2 & 37.4 $\pm$ 5.4 & \textbf{54.1 $\pm$ 4.9} \\
Top-5 (\%) & 72.1 $\pm$ 7.2 & 72.3 $\pm$ 7.0 & 71.3 $\pm$ 6.5 & \textbf{83.4 $\pm$ 3.4} \\
\bottomrule
\end{tabular}
}
\vspace{-1em}
\end{wraptable}

\textbf{LLM-based Text Generation.}
We compare two multimodal LLMs for generating per-image descriptions: 
LLaVA-1.5-7B~\cite{liu2023visual} and 
Qwen2-VL-7B~\cite{wang2024qwen2} (representative outputs in 
Fig.~\ref{fig:llm and prompt}, Appendix~\ref{app:supplementary_visualizations}). At matched $\alpha{=}0.1$, 
Qwen2-VL descriptions yield $54.05\%$ Top-1 vs.\ $53.02\%$ for LLaVA 
(Table~\ref{tab:alpha}), and both surpass the text-free baseline 
($\alpha{=}0$, $52.80\%$). The richer, more detailed descriptions from 
Qwen2-VL provide stronger semantic supervision, suggesting that 
text-encoder quality directly shapes the discriminability of the learned 
EEG representations.

\begin{wraptable}{r}{0.6\textwidth}
\vspace{-1em}
\centering
\small
\setlength{\tabcolsep}{4pt}
\renewcommand{\arraystretch}{1.12}
\caption{\textbf{Image retrieval performance} (mean $\pm$ std, \%) for different $\alpha$ values and models.}
\label{tab:alpha}
\resizebox{0.6\textwidth}{!}{
\begin{tabular}{lccc}
\toprule
Method / $\alpha$ & Top-1 & Top-3 & Top-5 \\
\midrule
$\alpha=0$ & 52.80 $\pm$ 7.85 & 74.02 $\pm$ 6.31 & 81.87 $\pm$ 5.79 \\
\midrule
LLaVA, $\alpha=0.1$ & 53.02 $\pm$ 7.31 & 74.03 $\pm$ 5.66 & 82.05 $\pm$ 5.29 \\
\midrule
\textbf{Qwen, \bm{$\alpha$} = 0.1} & \textbf{54.05 $\pm$ 4.87} & \textbf{75.73 $\pm$ 4.37} & \textbf{83.37 $\pm$ 3.39} \\
Qwen, $\alpha=0.3$ & 53.18 $\pm$ 6.53 & 74.60 $\pm$ 5.53 & 83.23 $\pm$ 4.47 \\
Qwen, $\alpha=0.5$ & 49.93 $\pm$ 6.64 & 71.92 $\pm$ 5.62 & 80.68 $\pm$ 4.79 \\
\bottomrule
\end{tabular}
}
\vspace{-1em}
\end{wraptable}

\textbf{Effect of the Alignment Weight $\alpha$}: The hyperparameter $\alpha \in [0,1]$ in Eq.~\ref{eq:total_loss} controls 
the weight of image--text supervision relative to EEG--image supervision. We sweep $\alpha$ and compare two LLMs on the EEG dataset
(Table~\ref{tab:alpha}). The best Top-1 accuracy is obtained at 
$\alpha{=}0.1$ with Qwen2-VL ($54.05\%$); performance remains close at 
$\alpha{=}0.3$ ($53.18\%$) but drops sharply at $\alpha{=}0.5$ 
($49.93\%$), falling below the text-free baseline 
($\alpha{=}0$, $52.80\%$). This pattern indicates that a small but non-zero $\alpha$ acts as a 
semantic regularizer: it injects linguistic structure into the shared 
embedding space without overwhelming the EEG--image objective. As larger $\alpha$ values 
favor text retrieval at the cost of image retrieval (Table~\ref{tab:text_retrieval_alpha}
, Appendix~\ref{app:additional_results}), we adopt $\alpha{=}0.1$ as default to align with prior image retrieval benchmark.

\vspace{-0.5em}
\begin{figure}[!h]
    \centering
    \includegraphics[width=0.9\linewidth]{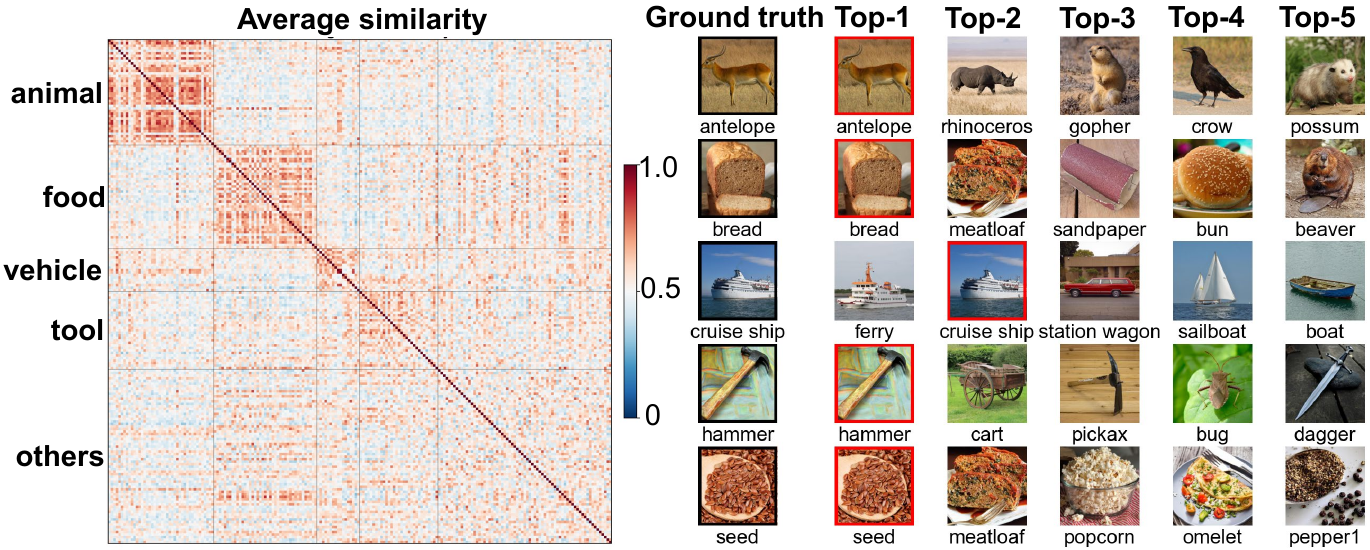}
    \vspace{-0.1em}
    \caption{\textbf{Semantic structure of learned EEG representations.} 
\textit{Left:} cosine-similarity matrix of EEG embeddings over 200 test concepts (averaged across 10 subjects); the 
block-diagonal pattern reveals intra-category clustering. \textit{Right:} 
top-5 image retrievals per category (correct matches in red); near-miss 
errors (e.g., \textit{cruise ship} $\rightarrow$ \textit{ferry}) reflect category-level proximity.}
    \vspace{-1.8em}
    \label{fig:similarity}
\end{figure}

\vspace{-0.8em}
\subsection{Semantic Analysis and Neural Dynamics}

\textbf{Semantic structure:} We perform representational similarity analysis 
(RSA)~\cite{kriegeskorte2008rsa,cichy2020eeg} on the learned EEG 
features (Fig.~\ref{fig:similarity}, left), grouping the 200 test 
concepts into five categories: animal, food, vehicle, tool, and others. 
Block-diagonal structure emerges, with intra-category similarity 
visibly exceeding inter-category similarity, most strongly for animals 
and food. This indicates that EEG representations encode category-level 
semantics despite training without category labels (subject-wise 
matrices in Fig.~\ref{fig:10subsimilarity}, 
Appendix~\ref{app:supplementary_visualizations}). The qualitative 
retrievals (Fig.~\ref{fig:similarity}, right) corroborate this: top-5 
candidates consistently fall within the ground-truth category, and 
near-miss errors are semantically adjacent (e.g., \textit{cruise ship} 
$\rightarrow$ \textit{ferry}), suggesting that decoding errors reflect coherent 
semantic proximity in the learned embedding space rather than noise.

\begin{wrapfigure}{r}{0.7\textwidth}
\centering
\includegraphics[width=\linewidth]{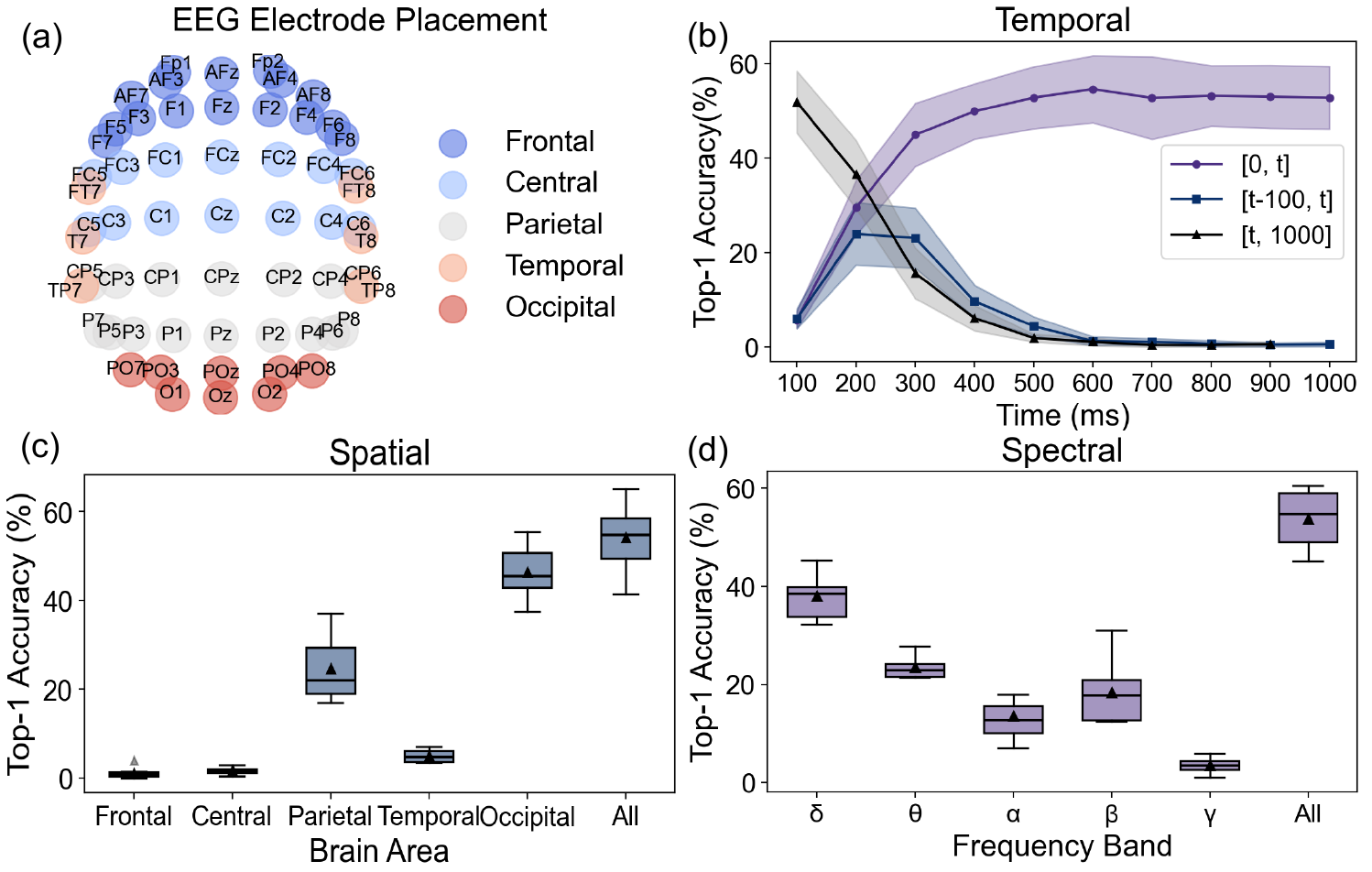}
\vspace{-1.5em}
\caption{\textbf{Temporal, spatial, and spectral analyses on EEG.} 
\textbf{(a)} Electrode layout, color-coded by anatomical region. 
\textbf{(b)} Spatial decoding by region: occipital sensors dominate, 
followed by temporal and parietal. 
\textbf{(c)} Temporal decoding under cumulative $[0, t]$, sliding $[t{-}100, t]$, and post-onset $[t, 1000]$~ms windows. 
\textbf{(d)} Spectral decoding across $\delta$, $\theta$, $\alpha$, $\beta$, $\gamma$, and full-band.}
\label{fig:electrode}
\vspace{-1.2em}
\end{wrapfigure}



\textbf{Temporal, spatial, and spectral dynamics.}
To assess biological plausibility, we examine where decoding information 
resides in time, space, and frequency in Fig.~\ref{fig:electrode}.
\textbf{\emph{Temporal}} (see Fig.~\ref{fig:electrode} (b)): the cumulative window $[0, 500]$~ms already achieves 
near-maximum Top-1 accuracy, while extending to $[0, 1000]$~ms yields 
only marginal gains and post-onset windows $[t, 1000]$~ms degrade 
sharply after $t{=}300$~ms; no single 100~ms sliding window matches the 
cumulative result, indicating that decoding integrates evidence 
distributed across the early window rather than relying on a single 
peak~\cite{cichy2014resolving}. 
\textbf{\emph{Spatial}} (see Fig.~\ref{fig:electrode} (c)): grouping electrodes  (see Fig.~\ref{fig:electrode} (a)) by 
anatomical region, occipital sensors contribute most strongly, followed 
by the temporal and parietal regions, while frontal and central electrodes 
contribute little—consistent with the role of the occipital cortex in early 
visual processing. 
\textbf{\emph{Spectral}} (see Fig.~\ref{fig:electrode} (d)): the delta band ($0.5$--$4$~Hz) 
yields the highest accuracy, with progressively weaker contributions 
from theta, alpha, beta, and gamma bands; this dominance likely reflects 
slow, time-locked event-related potentials rather than genuine delta 
oscillations~\cite{harmony2013functional}. 
These patterns 
align with the established neurophysiology of visual object recognition, 
indicating that the model relies on stimulus-driven brain activity rather than artifactual regularities.

\section{Conclusions, Limitations, and Future Work}

We present a tri-modal contrastive framework for EEG-based visual 
decoding that aligns noisy neural signals with visual and linguistic 
representations in a unified semantic space. By pre-training the EEG 
encoder with a masked reconstruction objective and aligning EEG, image, 
and LLM-generated text embeddings through contrastive learning, our 
method achieves substantial gains in decoding accuracy, cross-subject 
robustness, and semantic interpretability, demonstrating how 
self-supervised pre-training and language guidance can mitigate the 
weak supervision that has limited EEG-based decoding. The main 
limitation is that cross-subject Top-1 accuracy ($\sim$15\%) remains 
well below the in-subject ceiling, indicating that inter-subject 
variability is still unsolved. Our future work focuses on extending the proposed framework to MEG, fMRI, and generative reconstruction tasks such as diffusion-based image synthesis, opening a pathway toward 
semantically-grounded neural decoding for BCIs and assistive 
technologies.

\section*{Acknowledgments}
The project was partially funded by the Swedish Research Council (Vetenskapsrådet) under award 2023-00493, and the NAISS under award 2025/22-1173, 2025/23-185, and 2026/3-376. For the purpose of open access, the author has applied a Creative Commons Attribution (CC BY) license to any Author Accepted Manuscript version arising from this submission.



\section*{Reproducibility Statement}

We have made substantial efforts to ensure the reproducibility of our work. The paper provides detailed descriptions of the model architecture (Sec.~\ref{sec:methods}), training setup (Subsec.~\ref{sec:exp}), and ablation studies (Subsec.~\ref{sec:ablation}). Additional hyperparameters and implementation details are included in the Appendix. All datasets used in this work (Things-EEG2 and Things-MEG) are publicly available, and we describe the dataset preprocessing procedures in \autoref{app:preprocessing}. The source code and configuration files have already been publicly released on GitHub to facilitate full reproducibility. These instructions apply to everyone, regardless of the formatter being used.

\section*{Use of Large Language Models (LLMs)}
We used LLMs (e.g., ChatGPT and Claude) to rephrase and polish the manuscript and to assist with coding tasks. All LLM-generated code was reviewed, edited, and integrated by the authors; the LLM did not design algorithms or produce experimental results.

\clearpage

\bibliography{example_paper}

@article{miyawaki2008visual,
  title={Visual image reconstruction from human brain activity using a combination of multiscale local image decoders},
  author={Miyawaki, Yoichi and Uchida, Hajime and Yamashita, Okito and Sato, Masa-aki and Morito, Yusuke and Tanabe, Hiroki C and Sadato, Norihiro and Kamitani, Yukiyasu},
  journal={Neuron},
  volume={60},
  number={5},
  pages={915--929},
  year={2008},
  publisher={Elsevier}
}

@article{kay2008identifying,
  title={Identifying natural images from human brain activity},
  author={Kay, Kendrick N and Naselaris, Thomas and Prenger, Ryan J and Gallant, Jack L},
  journal={Nature},
  volume={452},
  number={7185},
  pages={352--355},
  year={2008},
  publisher={Nature Publishing Group UK London}
}

@article{puce2017review,
  title={A review of issues related to data acquisition and analysis in EEG/MEG studies},
  author={Puce, Aina and H{\"a}m{\"a}l{\"a}inen, Matti S},
  journal={Brain sciences},
  volume={7},
  number={6},
  pages={58},
  year={2017},
  publisher={MDPI}
}

@inproceedings{spampinato2017deep,
  title={Deep learning human mind for automated visual classification},
  author={Spampinato, Concetto and Palazzo, Simone and Kavasidis, Isaak and Giordano, Daniela and Souly, Nasim and Shah, Mubarak},
  booktitle={Proceedings of the IEEE conference on computer vision and pattern recognition},
  pages={6809--6817},
  year={2017}
}

@article{amin2017classification,
  title={Classification of EEG signals based on pattern recognition approach},
  author={Amin, Hafeez Ullah and Mumtaz, Wajid and Subhani, Ahmad Rauf and Saad, Mohamad Naufal Mohamad and Malik, Aamir Saeed},
  journal={Frontiers in computational neuroscience},
  volume={11},
  pages={103},
  year={2017},
  publisher={Frontiers Media SA}
}

@inproceedings{demir2021eeg,
  title={EEG-GNN: Graph neural networks for classification of electroencephalogram (EEG) signals},
  author={Demir, Andac and Koike-Akino, Toshiaki and Wang, Ye and Haruna, Masaki and Erdogmus, Deniz},
  booktitle={2021 43rd Annual International Conference of the IEEE Engineering in Medicine \& Biology Society (EMBC)},
  pages={1061--1067},
  year={2021},
  organization={IEEE}
}

@inproceedings{radford2021learning,
  title={Learning transferable visual models from natural language supervision},
  author={Radford, Alec and Kim, Jong Wook and Hallacy, Chris and Ramesh, Aditya and Goh, Gabriel and Agarwal, Sandhini and Sastry, Girish and Askell, Amanda and Mishkin, Pamela and Clark, Jack and others},
  booktitle={International conference on machine learning},
  pages={8748--8763},
  year={2021},
  organization={PmLR}
}

@article{wang2024qwen2,
  title={Qwen2-vl: Enhancing vision-language model's perception of the world at any resolution},
  author={Wang, Peng and Bai, Shuai and Tan, Sinan and Wang, Shijie and Fan, Zhihao and Bai, Jinze and Chen, Keqin and Liu, Xuejing and Wang, Jialin and Ge, Wenbin and others},
  journal={arXiv preprint arXiv:2409.12191},
  year={2024}
}

@article{dosovitskiy2020image,
  title={An image is worth 16x16 words: Transformers for image recognition at scale},
  author={Dosovitskiy, Alexey and Beyer, Lucas and Kolesnikov, Alexander and Weissenborn, Dirk and Zhai, Xiaohua and Unterthiner, Thomas and Dehghani, Mostafa and Minderer, Matthias and Heigold, Georg and Gelly, Sylvain and others},
  journal={arXiv preprint arXiv:2010.11929},
  year={2020}
}

@article{song2023decoding,
  author={Y. Song et al.},
  title={Decoding natural images from EEG for object recognition},
  journal={arXiv preprint arXiv:2308.13234},
  year={2023}
}

@article{hebart2023thingsdata,
  author  = {Martin N. Hebart and Oliver Contier and Lina Teichmann and Adam H. Rockter and Charles Y. Zheng and Alexis Kidder and Anna Corriveau and Maryam Vaziri-Pashkam and Chris I. Baker},
  title   = {THINGS-data, a multimodal collection of large-scale datasets for investigating object representations in human brain and behavior},
  journal = {eLife},
  volume  = {12},
  pages   = {e82580},
  year    = {2023},
  doi     = {10.7554/eLife.82580}
}

@inproceedings{wu2025bridging,
  title={Bridging the vision-brain gap with an uncertainty-aware blur prior},
  author={Wu, Haitao and Li, Qing and Zhang, Changqing and He, Zhen and Ying, Xiaomin},
  booktitle={Proceedings of the IEEE/CVF Conference on Computer Vision and Pattern Recognition},
  pages={2246--2257},
  year={2025}
}

@article{seeck2017standardized,
  title={The standardized EEG electrode array of the IFCN},
  author={Seeck, Margitta and Koessler, Laurent and Bast, Thomas and Leijten, Frans and Michel, Christoph and Baumgartner, Christoph and He, Bin and Beniczky, S{\'a}ndor},
  journal={Clinical neurophysiology},
  volume={128},
  number={10},
  pages={2070--2077},
  year={2017},
  publisher={Elsevier}
}

@article{kriegeskorte2008rsa,
  author={N. Kriegeskorte and M. Mur and P. Bandettini},
  title={Representational similarity analysis—connecting the branches of systems neuroscience},
  journal={Frontiers in Systems Neuroscience},
  volume={2},
  pages={4},
  year={2008}
}

@article{cichy2020eeg,
  title={AM/EEG-fMRI fusion primer: resolving human brain responses in space and time},
  author={Cichy, Radoslaw M and Oliva, Aude},
  journal={Neuron},
  volume={107},
  number={5},
  pages={772--781},
  year={2020},
  publisher={Elsevier}
}

@inproceedings{zhai2022scaling,
  title={Scaling vision transformers},
  author={Zhai, Xiaohua and Kolesnikov, Alexander and Houlsby, Neil and Beyer, Lucas},
  booktitle={Proceedings of the IEEE/CVF conference on computer vision and pattern recognition},
  pages={12104--12113},
  year={2022}
}

@article{yang2022chinese,
  title={Chinese clip: Contrastive vision-language pretraining in chinese},
  author={Yang, An and Pan, Junshu and Lin, Junyang and Men, Rui and Zhang, Yichang and Zhou, Jingren and Zhou, Chang},
  journal={arXiv preprint arXiv:2211.01335},
  year={2022}
}

@article{harmony2013functional,
  title={The functional significance of delta oscillations in cognitive processing},
  author={Harmony, Thal{\'\i}a},
  journal={Frontiers in integrative neuroscience},
  volume={7},
  pages={83},
  year={2013},
  publisher={Frontiers Media SA}
}

@article{cichy2014resolving,
  title={Resolving human object recognition in space and time},
  author={Cichy, Radoslaw Martin and Pantazis, Dimitrios and Oliva, Aude},
  journal={Nature neuroscience},
  volume={17},
  number={3},
  pages={455--462},
  year={2014},
  publisher={Nature Publishing Group US New York}
}

@article{loshchilov2017decoupled,
  title={Decoupled weight decay regularization},
  author={Loshchilov, Ilya and Hutter, Frank},
  journal={arXiv preprint arXiv:1711.05101},
  year={2017}
}

@article{kamitani2005decoding,
  title={Decoding the visual and subjective contents of the human brain},
  author={Kamitani, Yukiyasu and Tong, Frank},
  journal={Nature neuroscience},
  volume={8},
  number={5},
  pages={679--685},
  year={2005},
  publisher={Nature Publishing Group US New York}
}

@inproceedings{liu2024improved,
  title={Improved baselines with visual instruction tuning},
  author={Liu, Haotian and Li, Chunyuan and Li, Yuheng and Lee, Yong Jae},
  booktitle={Proceedings of the IEEE/CVF conference on computer vision and pattern recognition},
  pages={26296--26306},
  year={2024}
}

@article{grootswagers2022human,
  title={Human EEG recordings for 1,854 concepts presented in rapid serial visual presentation streams},
  author={Grootswagers, Tijl and Zhou, Ivy and Robinson, Amanda K and Hebart, Martin N and Carlson, Thomas A},
  journal={Scientific Data},
  volume={9},
  number={1},
  pages={3},
  year={2022},
  publisher={Nature Publishing Group UK London}
}

@article{li2024visual,
  title={Visual decoding and reconstruction via eeg embeddings with guided diffusion},
  author={Li, Dongyang and Wei, Chen and Li, Shiying and Zou, Jiachen and Qin, Haoyang and Liu, Quanying},
  journal={arXiv preprint arXiv:2403.07721},
  year={2024}
}

@article{song2025recognizing,
  title={Recognizing natural images from eeg with language-guided contrastive learning},
  author={Song, Yonghao and Wang, Yijun and He, Huiguang and Gao, Xiaorong},
  journal={IEEE Transactions on Neural Networks and Learning Systems},
  year={2025},
  publisher={IEEE}
}

@article{darvas2004mapping,
  title={Mapping human brain function with MEG and EEG: methods and validation},
  author={Darvas, Felix and Pantazis, D and Kucukaltun-Yildirim, E and Leahy, RM},
  journal={NeuroImage},
  volume={23},
  pages={S289--S299},
  year={2004},
  publisher={Elsevier}
}

@inproceedings{li2025neural,
  title={Neural-MCRL: Neural multimodal contrastive representation learning for EEG-based visual decoding},
  author={Li, Yueyang and Kang, Zijian and Gong, Shengyu and Dong, Wenhao and Zeng, Weiming and Yan, Hongjie and Siok, Wai Ting and Wang, Nizhuan},
  booktitle={2025 IEEE International Conference on Multimedia and Expo (ICME)},
  pages={1--6},
  year={2025},
  organization={IEEE}
}

@article{du2023decoding,
  title={Decoding visual neural representations by multimodal learning of brain-visual-linguistic features},
  author={Du, Changde and Fu, Kaicheng and Li, Jinpeng and He, Huiguang},
  journal={IEEE Transactions on Pattern Analysis and Machine Intelligence},
  volume={45},
  number={9},
  pages={10760--10777},
  year={2023},
  publisher={IEEE}
}

@article{scotti2024mindeye2,
  title={Mindeye2: Shared-subject models enable fmri-to-image with 1 hour of data},
  author={Scotti, Paul S and Tripathy, Mihir and Villanueva, Cesar Kadir Torrico and Kneeland, Reese and Chen, Tong and Narang, Ashutosh and Santhirasegaran, Charan and Xu, Jonathan and Naselaris, Thomas and Norman, Kenneth A and others},
  journal={arXiv preprint arXiv:2403.11207},
  year={2024}
}

@inproceedings{wang2024mindbridge,
  title={Mindbridge: A cross-subject brain decoding framework},
  author={Wang, Shizun and Liu, Songhua and Tan, Zhenxiong and Wang, Xinchao},
  booktitle={Proceedings of the IEEE/CVF Conference on Computer Vision and Pattern Recognition},
  pages={11333--11342},
  year={2024}
}

@article{liu2023visual,
  title={Visual instruction tuning},
  author={Liu, Haotian and Li, Chunyuan and Wu, Qingyang and Lee, Yong Jae},
  journal={Advances in neural information processing systems},
  volume={36},
  pages={34892--34916},
  year={2023}
}

@article{adeli2023predicting,
  title={Predicting brain activity using Transformers},
  author={Adeli, Hossein and Minni, Sun and Kriegeskorte, Nikolaus},
  journal={bioRxiv},
  pages={2023--08},
  year={2023},
  publisher={Cold Spring Harbor Laboratory}
}

@article{beliy2024wisdom,
  title={The wisdom of a crowd of brains: A universal brain encoder},
  author={Beliy, Roman and Wasserman, Navve and Zalcher, Amit and Irani, Michal},
  journal={arXiv preprint arXiv:2406.12179},
  year={2024}
}

@article{schirrmeister2017deep,
  title={Deep learning with convolutional neural networks for EEG decoding and visualization},
  author={Schirrmeister, Robin Tibor and Springenberg, Jost Tobias and Fiederer, Lukas Dominique Josef and Glasstetter, Martin and Eggensperger, Katharina and Tangermann, Michael and Hutter, Frank and Burgard, Wolfram and Ball, Tonio},
  journal={Human brain mapping},
  volume={38},
  number={11},
  pages={5391--5420},
  year={2017},
  publisher={Wiley Online Library}
}

@article{wang2018lstm,
  title={LSTM-based EEG classification in motor imagery tasks},
  author={Wang, Ping and Jiang, Aimin and Liu, Xiaofeng and Shang, Jing and Zhang, Li},
  journal={IEEE transactions on neural systems and rehabilitation engineering},
  volume={26},
  number={11},
  pages={2086--2095},
  year={2018},
  publisher={IEEE}
}

@article{zhong2020eeg,
  title={EEG-based emotion recognition using regularized graph neural networks},
  author={Zhong, Peixiang and Wang, Di and Miao, Chunyan},
  journal={IEEE Transactions on Affective Computing},
  volume={13},
  number={3},
  pages={1290--1301},
  year={2020},
  publisher={IEEE}
}

@article{velivckovic2017graph,
  title={Graph attention networks},
  author={Veli{\v{c}}kovi{\'c}, Petar and Cucurull, Guillem and Casanova, Arantxa and Romero, Adriana and Lio, Pietro and Bengio, Yoshua},
  journal={arXiv preprint arXiv:1710.10903},
  year={2017}
}

@article{brody2021attentive,
  title={How attentive are graph attention networks?},
  author={Brody, Shaked and Alon, Uri and Yahav, Eran},
  journal={arXiv preprint arXiv:2105.14491},
  year={2021}
}

@article{benchetrit2023brain,
  title={Brain decoding: toward real-time reconstruction of visual perception},
  author={Benchetrit, Yohann and Banville, Hubert and King, Jean-R{\'e}mi},
  journal={arXiv preprint arXiv:2310.19812},
  year={2023}
}

@inproceedings{guo2025neuro,
  title={Neuro-3d: Towards 3d visual decoding from eeg signals},
  author={Guo, Zhanqiang and Wu, Jiamin and Song, Yonghao and Bu, Jiahui and Mai, Weijian and Zheng, Qihao and Ouyang, Wanli and Song, Chunfeng},
  booktitle={Proceedings of the Computer Vision and Pattern Recognition Conference},
  pages={23870--23880},
  year={2025}
}

@inproceedings{he2022masked,
  title={Masked autoencoders are scalable vision learners},
  author={He, Kaiming and Chen, Xinlei and Xie, Saining and Li, Yanghao and Doll{\'a}r, Piotr and Girshick, Ross},
  booktitle={Proceedings of the IEEE/CVF conference on computer vision and pattern recognition},
  pages={16000--16009},
  year={2022}
}

@inproceedings{devlin2019bert,
  title={Bert: Pre-training of deep bidirectional transformers for language understanding},
  author={Devlin, Jacob and Chang, Ming-Wei and Lee, Kenton and Toutanova, Kristina},
  booktitle={Proceedings of the 2019 conference of the North American chapter of the association for computational linguistics: human language technologies, volume 1 (long and short papers)},
  pages={4171--4186},
  year={2019}
}

@article{hospedales2021meta,
  title={Meta-learning in neural networks: A survey},
  author={Hospedales, Timothy and Antoniou, Antreas and Micaelli, Paul and Storkey, Amos},
  journal={IEEE transactions on pattern analysis and machine intelligence},
  volume={44},
  number={9},
  pages={5149--5169},
  year={2021},
  publisher={IEEE}
}

@article{chien2022maeeg,
  title={Maeeg: Masked auto-encoder for eeg representation learning},
  author={Chien, Hsiang-Yun Sherry and Goh, Hanlin and Sandino, Christopher M and Cheng, Joseph Y},
  journal={arXiv preprint arXiv:2211.02625},
  year={2022}
}

@inproceedings{bai2024dreamdiffusion,
  title={DreamDiffusion: High-quality EEG-to-image generation with temporal masked signal modeling and CLIP alignment},
  author={Bai, Yunpeng and Wang, Xintao and Cao, Yan-Pei and Ge, Yixiao and Yuan, Chun and Shan, Ying},
  booktitle={European Conference on Computer Vision},
  pages={472--488},
  year={2024},
  organization={Springer}
}

@article{wang2025eegmamba,
  title={Eegmamba: An eeg foundation model with mamba},
  author={Wang, Jiquan and Zhao, Sha and Luo, Zhiling and Zhou, Yangxuan and Li, Shijian and Pan, Gang},
  journal={Neural Networks},
  pages={107816},
  year={2025},
  publisher={Elsevier}
}

@article{mathis2024decoding,
  title={Decoding the brain: From neural representations to mechanistic models},
  author={Mathis, Mackenzie Weygandt and Rotondo, Adriana Perez and Chang, Edward F and Tolias, Andreas S and Mathis, Alexander},
  journal={Cell},
  volume={187},
  number={21},
  pages={5814--5832},
  year={2024},
  publisher={Elsevier}
}

@article{yamins2016using,
  title={Using goal-driven deep learning models to understand sensory cortex},
  author={Yamins, Daniel LK and DiCarlo, James J},
  journal={Nature neuroscience},
  volume={19},
  number={3},
  pages={356--365},
  year={2016},
  publisher={Nature Publishing Group}
}

@article{caucheteux2022brains,
  title={Brains and algorithms partially converge in natural language processing},
  author={Caucheteux, Charlotte and King, Jean-R{\'e}mi},
  journal={Communications biology},
  volume={5},
  number={1},
  pages={134},
  year={2022},
  publisher={Nature Publishing Group UK London}
}

@article{haxby2011common,
  title={A common, high-dimensional model of the representational space in human ventral temporal cortex},
  author={Haxby, James V and Guntupalli, J Swaroop and Connolly, Andrew C and Halchenko, Yaroslav O and Conroy, Bryan R and Gobbini, M Ida and Hanke, Michael and Ramadge, Peter J},
  journal={Neuron},
  volume={72},
  number={2},
  pages={404--416},
  year={2011},
  publisher={Elsevier}
}

@article{naselaris2011encoding,
  title={Encoding and decoding in fMRI},
  author={Naselaris, Thomas and Kay, Kendrick N and Nishimoto, Shinji and Gallant, Jack L},
  journal={Neuroimage},
  volume={56},
  number={2},
  pages={400--410},
  year={2011},
  publisher={Elsevier}
}

@article{huth2016natural,
  title={Natural speech reveals the semantic maps that tile human cerebral cortex},
  author={Huth, Alexander G and De Heer, Wendy A and Griffiths, Thomas L and Theunissen, Fr{\'e}d{\'e}ric E and Gallant, Jack L},
  journal={Nature},
  volume={532},
  number={7600},
  pages={453--458},
  year={2016},
  publisher={Nature Publishing Group UK London}
}

@article{yamins2014performance,
  title={Performance-optimized hierarchical models predict neural responses in higher visual cortex},
  author={Yamins, Daniel LK and Hong, Ha and Cadieu, Charles F and Solomon, Ethan A and Seibert, Darren and DiCarlo, James J},
  journal={Proceedings of the national academy of sciences},
  volume={111},
  number={23},
  pages={8619--8624},
  year={2014},
  publisher={National Academy of Sciences}
}

@article{kriegeskorte2015deep,
  title={Deep neural networks: a new framework for modeling biological vision and brain information processing},
  author={Kriegeskorte, Nikolaus},
  journal={Annual review of vision science},
  volume={1},
  pages={417--446},
  year={2015},
  publisher={Annual Reviews}
}

@article{d2025tribe,
  title={TRIBE: TRImodal Brain Encoder for whole-brain fMRI response prediction},
  author={d'Ascoli, St{\'e}phane and Rapin, J{\'e}r{\'e}my and Benchetrit, Yohann and Banville, Hubert and King, Jean-R{\'e}mi},
  journal={arXiv preprint arXiv:2507.22229},
  year={2025}
}

@article{schneider2023learnable,
  title={Learnable latent embeddings for joint behavioural and neural analysis},
  author={Schneider, Steffen and Lee, Jin Hwa and Mathis, Mackenzie Weygandt},
  journal={Nature},
  volume={617},
  number={7960},
  pages={360--368},
  year={2023},
  publisher={Nature Publishing Group UK London}
}

@article{wang2025foundation,
  title={Foundation model of neural activity predicts response to new stimulus types},
  author={Wang, Eric Y and Fahey, Paul G and Ding, Zhuokun and Papadopoulos, Stelios and Ponder, Kayla and Weis, Marissa A and Chang, Andersen and Muhammad, Taliah and Patel, Saumil and Ding, Zhiwei and others},
  journal={Nature},
  volume={640},
  number={8058},
  pages={470--477},
  year={2025},
  publisher={Nature Publishing Group UK London}
}

@article{binz2025foundation,
  title={A foundation model to predict and capture human cognition},
  author={Binz, Marcel and Akata, Elif and Bethge, Matthias and Br{\"a}ndle, Franziska and Callaway, Fred and Coda-Forno, Julian and Dayan, Peter and Demircan, Can and Eckstein, Maria K and {\'E}ltet{\H{o}}, No{\'e}mi and others},
  journal={Nature},
  volume={644},
  number={8078},
  pages={1002--1009},
  year={2025},
  publisher={Nature Publishing Group UK London}
}

@article{schrimpf2018brain,
  title={Brain-score: Which artificial neural network for object recognition is most brain-like?},
  author={Schrimpf, Martin and Kubilius, Jonas and Hong, Ha and Majaj, Najib J and Rajalingham, Rishi and Issa, Elias B and Kar, Kohitij and Bashivan, Pouya and Prescott-Roy, Jonathan and Geiger, Franziska and others},
  journal={BioRxiv},
  pages={407007},
  year={2018},
  publisher={Cold Spring Harbor Laboratory}
}

@article{gifford2022large,
  title={A large and rich EEG dataset for modeling human visual object recognition},
  author={Gifford, Alessandro T and Dwivedi, Kshitij and Roig, Gemma and Cichy, Radoslaw M},
  journal={NeuroImage},
  volume={264},
  pages={119754},
  year={2022},
  publisher={Elsevier}
}

@article{ouahidi2025reve,
  title={REVE: A Foundation Model for EEG--Adapting to Any Setup with Large-Scale Pretraining on 25,000 Subjects},
  author={Ouahidi, Yassine El and Lys, Jonathan and Th{\"o}lke, Philipp and Farrugia, Nicolas and Pasdeloup, Bastien and Gripon, Vincent and Jerbi, Karim and Lioi, Giulia},
  journal={arXiv preprint arXiv:2510.21585},
  year={2025}
}

@article{zhou2025spiced,
  title={SPICED: A Synaptic Homeostasis-Inspired Framework for Unsupervised Continual EEG Decoding},
  author={Zhou, Yangxuan and Zhao, Sha and Wang, Jiquan and Jiang, Haiteng and Li, Shijian and Li, Tao and Pan, Gang},
  journal={arXiv preprint arXiv:2509.17439},
  year={2025}
}

@article{yang2025thd,
  title={THD-BAR: Topology Hierarchical Derived Brain Autoregressive Modeling for EEG Generic Representations},
  author={Yang, Wenchao and Yan, Weidong and Liu, Wenkang and Ma, Yulan and Li, Yang},
  journal={arXiv preprint arXiv:2511.13733},
  year={2025}
}

@inproceedings{kneeland2025enigma,
  title={ENIGMA: A Unified Lightweight EEG-to-Image Model for Multi-Subject Visual Decoding},
  author={Kneeland, Reese and Jiang, Wangshu and Nunes, Ugo Bruzadin and Lee, Si Kai and Scotti, Paul Steven and Delorme, Arnaud and Xu, Jonathan},
  booktitle={NeurIPS 2025 Workshop on Foundation Models for the Brain and Body},
  year={2025}
}

@article{fang2025neuript,
  title={Neuript: Foundation model for neural interfaces},
  author={Fang, Zitao and Li, Chenxuan and Zhou, Hongting and Yu, Shuyang and Du, Guodong and Qasem, Ashwaq and Lu, Yang and Li, Jing and Zhang, Junsong and Goh, Sim Kuan},
  journal={arXiv preprint arXiv:2510.16548},
  year={2025}
}
\bibliographystyle{unsrtnat}

\clearpage
\appendix
\onecolumn

\clearpage 

\appendix

\section{Dataset}
\label{app:dataset}

We evaluate our method on two large-scale benchmarks: Things-EEG2 and Things-MEG. Table~\ref{tab:datasets} provides the detailed information on the two datasets. Things-EEG2 provides 63-channel EEG recordings from 10 participants viewing natural object images under a rapid serial visual presentation (RSVP) paradigm with a 200~ms stimulus onset asynchrony (100~ms image + 100~ms blank). The training set spans 1{,}654 concepts (10 images $\times$ 4 repetitions each), and the test set contains 200 held-out concepts (1 image $\times$ 80 repetitions) strictly disjoint from training, forming a 200-way zero-shot retrieval protocol. 

Things-MEG provides MEG recordings from 4 participants viewing 1{,}854 concepts (12 images per concept; 22{,}248 images in total) from the THINGS stimulus set. Most images were presented once, while a repeated-image test set consisting of 200 images was presented 12 times across sessions for model evaluation and response reliability assessment.

\begin{table*}[!h]
\centering
\caption{Summary of the \textbf{Things-EEG2 and Things-MEG datasets} used in experiments.}
\label{tab:datasets}
\small
\begin{tabular}{l@{\hspace{1pt}}c@{\hspace{1pt}}c@{\hspace{1pt}}c@{\hspace{1pt}}c@{\hspace{1pt}}c@{\hspace{1pt}}c}
\toprule
\textbf{Data} & \textbf{Subject} & \textbf{Channel} & \textbf{Training Set} & \textbf{Testing Set} & \textbf{SOA} \\
\midrule
EEG & 10 & 63 & 1,654 concepts $\times$ 10 imgs $\times$ 4 reps & 200 concepts $\times$ 1 img $\times$ 80 reps & 200 ms \\
MEG & 4 & 271 & 1,854 concepts $\times$ 12 imgs $\times$ 1 rep & 200 concepts $\times$ 1 img $\times$ 12 reps & 1500 $\pm$ 200 ms \\
\bottomrule
\end{tabular}
\end{table*}

\section{Preprocessing and Implementation}
\label{app:preprocessing}

\textbf{Preprocessing.} EEG signals were processed using the public 
Things-EEG2 pipeline: re-referenced to the average of all electrodes, 
band-pass filtered to $0.1$--$100$~Hz, baseline-corrected to the 200~ms 
pre-stimulus window, downsampled to 250~Hz, epoched over $0$--$1000$~ms 
post-stimulus onset, and averaged across repetitions. No ICA, additional 
artifact rejection, or data augmentation, was applied.

For MEG signals in Things-MEG dataset, the data were band-pass filtered to $0.1$--$100$~Hz, 
downsampled to 250~Hz, and epoched over $0$--$1000$~ms relative to stimulus onset. 
Repeated-image trials were averaged across repetitions.

\textbf{Implementation.} The framework is implemented in PyTorch 
(Python~3.12) and trained on a single NVIDIA RTX~4090, requiring 
$\sim$5~mins per subject for Stage~1 and $\sim$3~mins for Stage~2. We 
optimize with AdamW~\cite{loshchilov2017decoupled} 
(lr $=2\times10^{-4}$, $\beta_1{=}0.5$, $\beta_2{=}0.999$); batch sizes 
are 1{,}000 for Things-EEG2 and 500 for Things-MEG. Stage~1 (MAE 
pre-training) runs 200 epochs with masking ratio 0.3 and decoder 
($W{=}256$, $D{=}2$); Stage~2 (alignment) runs up to 150 epochs with 
early stopping (patience 10) and $\alpha{=}0.1$. From the 16{,}540 
training trials, 740 are held out for validation, fixed across runs 
and seeds. Final predictions average the three checkpoints with the 
lowest validation loss; all experiments are repeated over 3 seeds.

\textbf{Statistical testing.} We assess significance with paired Wilcoxon 
signed-rank tests over the 10 per-subject scores (two-sided, 
$\alpha{=}0.05$), applying Holm correction across baselines. We report 
$p$-values and rank-biserial effect sizes, and interpret results 
conservatively given the small sample ($N{=}10$).

\textbf{Vision and text encoders.} We use publicly available pretrained CLIP models implemented in HuggingFace Transformers, and the details of the selected models are listed in Table~\ref{tab:vision_encoder_overview}.

\begin{table*}[!h]
\centering
\caption{\textbf{CLIP models} used as \textbf{vision and text encoders} in the experiments.}
\renewcommand{\arraystretch}{1.2}
\small
\begin{tabular}{l@{\hspace{1pt}}c@{\hspace{1pt}}c@{\hspace{1pt}}c@{\hspace{0pt}}c}
\hline
\textbf{Model} & \textbf{Params (M)} & \textbf{Training Data / Scale} & \textbf{Visual Backbone} & \textbf{Emb Dim} \\
\hline
ViT-L-14~\cite{dosovitskiy2020image} & 428 & OpenAI CLIP WebImageText corpus & ViT-L/14 & 768 \\
\hline
ViT-H-14~\cite{dosovitskiy2020image} & 986 & LAION-2B English subset (approx.~2B pairs) & ViT-H/14 & 1024 \\
\hline
ViT-G-14~\cite{zhai2022scaling} & 1370 & LAION-2B English subset (approx.~2B pairs) & ViT-G/14 & 1024 \\
\hline
CN‑CLIP~\cite{yang2022chinese} & 38 & Chinese WebImage‑Text (approx.~200M pairs) & ResNet50 & 1024 \\
\hline
\end{tabular}
\label{tab:vision_encoder_overview}
\end{table*}

\FloatBarrier
\section{Additional Experimental Results}
\label{app:additional_results}

We provide per-subject breakdowns and additional ablations on the EEG dataset, including Top-1 and Top-5 accuracies for all 10 subjects, including in-subject and cross-subject image retrieval results
(Tables~\ref{tab:accuracy_in_subject} and~\ref{tab:accuracy_cross_subject}), 
text retrieval across $\alpha$ values 
(Table~\ref{tab:text_retrieval_alpha}), per-subject results of different image encoders (Table~\ref{tab:image_encoder_subjects}, per-subject results of different EEG encoders 
(Table~\ref{tab:top1_top5_by_encoder}), EEG encoder ablation studies (Table~\ref{tab:eeg_ablation_study}), and the LLM prompt with example 
outputs generated using the Qwen2-VL-7B model (Table~\ref{tab:llm_prompt_examples}).

\begin{table*}[!htbp]
\centering
\caption{\textbf{Top-1 and Top-5 image retrieval accuracy (\%) in subjects}. (NICE, NICE++, ATMS, MCRL refer to results reported in the original paper)}
\label{tab:accuracy_in_subject}
\renewcommand{\arraystretch}{1.2}
\scriptsize
\setlength{\tabcolsep}{3pt}
\resizebox{\textwidth}{!}{
\begin{tabular}{lcccccccccccccccccccc}
\hline
\multirow{2}{*}{Model} 
& \multicolumn{2}{c}{Sub1} & \multicolumn{2}{c}{Sub2} & \multicolumn{2}{c}{Sub3} & \multicolumn{2}{c}{Sub4} & \multicolumn{2}{c}{Sub5}
& \multicolumn{2}{c}{Sub6} & \multicolumn{2}{c}{Sub7} & \multicolumn{2}{c}{Sub8} & \multicolumn{2}{c}{Sub9} & \multicolumn{2}{c}{Sub10} \\
\cmidrule(lr){2-3}\cmidrule(lr){4-5}\cmidrule(lr){6-7}\cmidrule(lr){8-9}\cmidrule(lr){10-11}
\cmidrule(lr){12-13}\cmidrule(lr){14-15}\cmidrule(lr){16-17}\cmidrule(lr){18-19}\cmidrule(lr){20-21}
& Top-1 & Top-5 & Top-1 & Top-5 & Top-1 & Top-5 & Top-1 & Top-5 & Top-1 & Top-5
& Top-1 & Top-5 & Top-1 & Top-5 & Top-1 & Top-5 & Top-1 & Top-5 & Top-1 & Top-5 \\
\hline

NICE~\cite{song2023decoding}   & 12.3 & 36.6 & 10.4 & 33.9 & 13.1 & 39.0 & 16.4 & 47.0 & 8.0  & 26.9
       & 14.1 & 40.6 & 15.2 & 42.1 & 20.0 & 49.9 & 13.3 & 37.1 & 14.9 & 41.9 \\

NICE++~\cite{song2025recognizing} & 14.5 & 41.8 & 16.7 & 43.4 & 18.2 & 47.3 & 21.1 & 54.8 & 14.2 & 38.7
       & 16.0 & 46.8 & 17.9 & 48.2 & 22.7 & 59.9 & 17.4 & 45.3 & 19.1 & 50.1 \\

ATMS~\cite{li2024visual}   & 21.0 & 51.5 & 24.5 & 54.0 & 27.0 & 61.0 & 18.5 & 49.5 & 29.5 & 44.5
       & 24.6 & 59.5 & 25.5 & 57.0 & 37.0 & 72.0 & 26.0 & 53.5 & 34.0 & 69.5 \\

MCRL~\cite{li2025neural}   & 27.5 & 64.0 & 28.5 & 61.5 & 37.0 & 69.0 & 35.0 & 66.0 & 22.5 & 51.5
       & 31.5 & 61.0 & 31.5 & 62.5 & 42.0 & 74.5 & 30.5 & 59.5 & 37.5 & 71.0 \\

\textbf{Ours} 
& \textbf{56.5} & \textbf{85.5} & \textbf{52.3} & \textbf{81.8} 
& \textbf{53.3} & \textbf{79.7} & \textbf{56.7} & \textbf{86.7} 
& \textbf{47.5} & \textbf{80.5}
& \textbf{50.3} & \textbf{83.3} & \textbf{50.1} & \textbf{80.3} 
& \textbf{64.0} & \textbf{87.0} & \textbf{51.5} & \textbf{80.0} 
& \textbf{58.3} & \textbf{88.8} \\

\hline
\end{tabular}
}
\end{table*}

\begin{table*}[!htbp]
\centering
\caption{\textbf{Top-1 and Top-5 image retrieval accuracy (\%) cross subjects.} (NICE, NICE++, ATMS, MCRL refer to results reported in the original paper)}
\label{tab:accuracy_cross_subject}
\renewcommand{\arraystretch}{1.2}
\scriptsize
\setlength{\tabcolsep}{3pt}
\resizebox{\textwidth}{!}{
\begin{tabular}{lcccccccccccccccccccc}
\hline
\multirow{2}{*}{Model}
& \multicolumn{2}{c}{Sub1} & \multicolumn{2}{c}{Sub2} & \multicolumn{2}{c}{Sub3} & \multicolumn{2}{c}{Sub4} & \multicolumn{2}{c}{Sub5}
& \multicolumn{2}{c}{Sub6} & \multicolumn{2}{c}{Sub7} & \multicolumn{2}{c}{Sub8} & \multicolumn{2}{c}{Sub9} & \multicolumn{2}{c}{Sub10} \\
\cmidrule(lr){2-3}\cmidrule(lr){4-5}\cmidrule(lr){6-7}\cmidrule(lr){8-9}\cmidrule(lr){10-11}
\cmidrule(lr){12-13}\cmidrule(lr){14-15}\cmidrule(lr){16-17}\cmidrule(lr){18-19}\cmidrule(lr){20-21}
& Top-1 & Top-5 & Top-1 & Top-5 & Top-1 & Top-5 & Top-1 & Top-5 & Top-1 & Top-5
& Top-1 & Top-5 & Top-1 & Top-5 & Top-1 & Top-5 & Top-1 & Top-5 & Top-1 & Top-5 \\
\hline

NICE~\cite{song2023decoding}   & 7.6  & 22.8 & 5.9  & 20.5 & 6.0  & 22.3 & 6.3  & 20.7 & 4.4  & 18.3
       & 5.6  & 22.2 & 5.6  & 19.7 & 6.3  & 22.0 & 5.7  & 17.6 & 8.4  & 28.3 \\

ATMS~\cite{li2024visual}   & 9.5  & 24.5 & 11.5 & 33.5 & 8.5  & 29.5 & 11.5 & 30.0 & 8.5  & 24.0
       & 10.5 & 27.5 & 8.0  & 26.5 & 13.5 & 30.5 & 9.5  & 27.5 & 12.5 & 37.0 \\

MCRL~\cite{li2025neural}   & 13.0 & 31.5 & 12.0 & 30.5 & \textbf{14.5} & \textbf{35.5} & 12.5 & 35.0 & \textbf{11.5} & 29.0
       & 13.5 & \textbf{35.5} & 14.0 & 36.0 & \textbf{18.5} & 38.5 & \textbf{13.5} & \textbf{32.5} & 17.0 & 39.0 \\

\textbf{Ours} 
& \textbf{16.3} & \textbf{46.2} & \textbf{21.8} & \textbf{48.7} 
& 13.3 & 35.3 & \textbf{15.0} & \textbf{35.8} 
& 11.2 & \textbf{35.0}
& \textbf{13.8} & 35.0 & \textbf{14.3} & \textbf{36.8} 
& 14.0 & \textbf{39.2} & 7.8  & 29.5
& \textbf{23.8} & \textbf{51.5} \\

\hline
\end{tabular}
}
\end{table*}

\begin{table*}[!htbp]
\centering
\caption{\textbf{Top-1 and Top-5 image retrieval accuracy (\%) across subjects for different vision backbones of CLIP model}.}
\label{tab:image_encoder_subjects}
\scriptsize
\setlength{\tabcolsep}{3pt}
\renewcommand{\arraystretch}{1.2}
\resizebox{\textwidth}{!}{
\begin{tabular}{lcccccccccccccccccccc}
\hline
\multirow{2}{*}{Model} 
& \multicolumn{2}{c}{Sub1} & \multicolumn{2}{c}{Sub2} & \multicolumn{2}{c}{Sub3} & \multicolumn{2}{c}{Sub4} & \multicolumn{2}{c}{Sub5}
& \multicolumn{2}{c}{Sub6} & \multicolumn{2}{c}{Sub7} & \multicolumn{2}{c}{Sub8} & \multicolumn{2}{c}{Sub9} & \multicolumn{2}{c}{Sub10} \\
\cmidrule(lr){2-3}\cmidrule(lr){4-5}\cmidrule(lr){6-7}\cmidrule(lr){8-9}\cmidrule(lr){10-11}\cmidrule(lr){12-13}\cmidrule(lr){14-15}\cmidrule(lr){16-17}\cmidrule(lr){18-19}\cmidrule(lr){20-21}
& Top-1 & Top-5 & Top-1 & Top-5 & Top-1 & Top-5 & Top-1 & Top-5 & Top-1 & Top-5
& Top-1 & Top-5 & Top-1 & Top-5 & Top-1 & Top-5 & Top-1 & Top-5 & Top-1 & Top-5 \\
\hline
ViT-L-14~\cite{dosovitskiy2020image} & 38.2 & 73.8 & 36.7 & 68.0 & 42.7 & 70.3 & 47.0 & 80.5 & 25.3 & 56.3 & 35.5 & 73.7 & 36.0 & 70.2 & 52.3 & 79.5 & 33.8 & 69.2 & 48.0 & 79.2 \\
ViT-H-14~\cite{dosovitskiy2020image} & 36.8 & 70.3 & 37.5 & 71.0 & 40.7 & 73.7 & 44.7 & 77.7 & 32.3 & 59.5 & 38.2 & 74.5 & 37.3 & 66.5 & 52.0 & 78.2 & 38.5 & 67.2 & 55.2 & 84.3 \\
ViT-G-14\cite{zhai2022scaling} & 33.3 & 68.8 & 33.0 & 68.3 & 40.5 & 72.2 & 40.7 & 78.5 & 31.7 & 61.3 & 36.8 & 70.3 & 37.5 & 71.0 & 45.8 & 78.2 & 36.2 & 68.3 & 45.0 & 78.0 \\
CN-CLIP~\cite{yang2022chinese} & \textbf{58.0} & \textbf{87.2} & \textbf{55.2} & \textbf{83.5} & \textbf{45.8} & \textbf{77.5} & \textbf{53.8} & \textbf{87.0} & \textbf{43.7} & \textbf{76.5} & \textbf{52.2} & \textbf{84.2} & \textbf{49.5} & \textbf{82.7} & \textbf{68.7} & \textbf{89.7} & \textbf{49.7} & \textbf{79.2} & \textbf{60.7} & \textbf{90.7} \\
\hline
\end{tabular}
}
\end{table*}

\begin{table*}[!htbp]
\centering
\scriptsize
\renewcommand{\arraystretch}{1.2}
\setlength{\tabcolsep}{3pt}
\caption{\textbf{Text retrieval accuracy (\%) across subjects for different $\alpha$ values.} Compared with Table~\ref{tab:alpha}, higher $\alpha$ values generally lead to better text retrieval but worse image retrieval performance.}
\label{tab:text_retrieval_alpha}
\resizebox{\textwidth}{!}{
\begin{tabular}{lcccccccccccccccccccccc}
\hline
\multirow{2}{*}{\makecell{Alpha\\($\alpha$)}} & \multicolumn{2}{c}{Sub1} & \multicolumn{2}{c}{Sub2} & \multicolumn{2}{c}{Sub3} & \multicolumn{2}{c}{Sub4} & \multicolumn{2}{c}{Sub5}
& \multicolumn{2}{c}{Sub6} & \multicolumn{2}{c}{Sub7} & \multicolumn{2}{c}{Sub8} & \multicolumn{2}{c}{Sub9} & \multicolumn{2}{c}{Sub10} & \multicolumn{2}{c}{Ave} \\
\cmidrule(lr){2-3}\cmidrule(lr){4-5}\cmidrule(lr){6-7}\cmidrule(lr){8-9}\cmidrule(lr){10-11}\cmidrule(lr){12-13}\cmidrule(lr){14-15}\cmidrule(lr){16-17}\cmidrule(lr){18-19}\cmidrule(lr){20-21}\cmidrule(lr){22-23}
& Top-1 & Top-5 & Top-1 & Top-5 & Top-1 & Top-5 & Top-1 & Top-5 & Top-1 & Top-5 & Top-1 & Top-5 & Top-1 & Top-5 & Top-1 & Top-5 & Top-1 & Top-5 & Top-1 & Top-5 & Top-1 & Top-5 \\
\hline
0.1 & 10.8 & 27.7 & 7.5 & 25.7 & 6.7 & 25.2 & 9.2 & 25.8 & 8.0 & 26.2 & 7.0 & 26.2 & 7.5 & 23.8 & 8.8 & 31.7 & 10.0 & 22.8 & 8.7 & 28.7 & 8.4 & 26.4 \\
0.2 & \textbf{12.7} & 29.2 & 8.8 & 26.5 & \textbf{10.3} & 29.0 & 11.0 & 30.5 & 10.5 & 26.2 & 8.0 & 28.5 & 10.7 & 30.5 & 12.2 & 33.2 & 12.0 & 26.7 & 9.5 & 30.8 & 10.6 & 29.1 \\
0.5 & 11.5 & 34.5 & 11.3 & 29.3 & 8.5 & 31.3 & 12.7 & 29.7 & 11.0 & \textbf{27.5} & 10.5 & 31.2 & \textbf{12.3} & \textbf{31.8} & \textbf{13.5} & 35.3 & 11.2 & \textbf{30.2} & 11.0 & 37.8 & 11.3 & 31.9 \\
0.7 & 12.0 & 34.7 & \textbf{11.7} & \textbf{29.7} & 9.3 & \textbf{32.7} & \textbf{14.0} & \textbf{30.7} & 10.7 & 26.8 & \textbf{10.7} & 33.3 & 12.2 & 31.5 & 13.3 & 34.3 & \textbf{12.3} & 29.2 & \textbf{11.5} & \textbf{38.8} & \textbf{11.8} & 32.2 \\
0.9 & 12.0 & \textbf{34.8} & 9.7 & \textbf{29.7} & 9.3 & 32.0 & 13.2 & 30.3 & \textbf{11.2} & 26.0 & 8.8 & \textbf{35.8} & \textbf{12.3} & 31.3 & 12.0 & \textbf{35.8} & 11.3 & 29.2 & 11.3 & 38.0 & 11.1 & \textbf{32.3} \\
\hline
\end{tabular}
}
\end{table*}

\begin{table*}[!htbp]
\centering
\scriptsize
\setlength{\tabcolsep}{3pt}
\renewcommand{\arraystretch}{1.2}
\caption{\textbf{Image retrieval accuracy (\%) across subjects for different EEG encoders}. NICE, ATMS, MCRL refer to the EEG encoders proposed in the corresponding original paper, which are re-implemented within our framework, under the same tri-modal alignment setting and using the same vision backbone (CN-CLIP) for fair comparison, but without applying our pre-training strategy.}
\label{tab:top1_top5_by_encoder}
\resizebox{\textwidth}{!}{
\begin{tabular}{lcccccccccccccccccccccc}
\toprule
\multirow{2}{*}{Encoder} 
& \multicolumn{2}{c}{Sub1} & \multicolumn{2}{c}{Sub2} & \multicolumn{2}{c}{Sub3}
& \multicolumn{2}{c}{Sub4} & \multicolumn{2}{c}{Sub5} & \multicolumn{2}{c}{Sub6}
& \multicolumn{2}{c}{Sub7} & \multicolumn{2}{c}{Sub8} & \multicolumn{2}{c}{Sub9}
& \multicolumn{2}{c}{Sub10} & \multicolumn{2}{c}{Ave} \\
\cmidrule(lr){2-3}\cmidrule(lr){4-5}\cmidrule(lr){6-7}\cmidrule(lr){8-9}\cmidrule(lr){10-11}\cmidrule(lr){12-13}\cmidrule(lr){14-15}\cmidrule(lr){16-17}\cmidrule(lr){18-19}\cmidrule(lr){20-21}\cmidrule(lr){22-23}
& Top-1 & Top-5 & Top-1 & Top-5 & Top-1 & Top-5 & Top-1 & Top-5 & Top-1 & Top-5
& Top-1 & Top-5 & Top-1 & Top-5 & Top-1 & Top-5 & Top-1 & Top-5 & Top-1 & Top-5 & Top-1 & Top-5 \\
\midrule
NICE~\cite{song2023decoding}    & 44.0 & 77.8 & 43.2 & 73.2 & 46.0 & \textbf{81.5} & 51.8 & 86.8 & 39.5 & 70.2 & 47.5 & 77.7 & 41.2 & 75.2 & 60.3 & 89.3 & 43.8 & 76.8 & 57.2 & 88.5 & 47.5 & 79.7 \\
ATMS~\cite{li2024visual}   & 56.7 & 86.0 & 51.3 & \textbf{82.2} & \textbf{50.0} & 80.7 & 55.2 & 87.0 & 41.0 & \textbf{76.8} & \textbf{51.7} & 81.3 & \textbf{49.5} & 81.2 & 67.0 & \textbf{90.0} & 45.2 & 80.3 & \textbf{64.2} & 90.3 & 53.2 & \textbf{83.6} \\
MCRL~\cite{li2025neural}   & \textbf{58.8} & \textbf{88.3} & \textbf{56.3} & 82.0 & 42.7 & 75.8 & \textbf{57.5} & \textbf{88.3} & 39.2 & 69.2 & 51.0 & 80.0 & 47.2 & \textbf{83.8} & 65.8 & 89.3 & \textbf{51.8} & \textbf{83.5} & 61.8 & \textbf{92.2} & 53.2 & 83.3 \\
\textbf{Ours}   & 55.3 & 84.5 & 53.2 & 82.0 & 45.2 & 79.5 & 51.7 & 86.5 & \textbf{42.5} & 71.5 & 50.2 & \textbf{84.5} & 49.0 & 82.2 & \textbf{68.3} & 87.8 & 50.2 & 78.7 & 59.8 & 90.7 & 52.5 & 82.8 \\
\midrule
\multicolumn{23}{l}{\textbf{Pretrained EEG Encoder}} \\
\midrule
\textbf{Ours}   & 56.5 & 85.5 & 52.3 & 81.8 & \textbf{53.3} & 79.7 & 56.7 & 86.7 & \textbf{47.5} & \textbf{80.5} & 50.3 & 83.3 & \textbf{50.0} & 80.3 & 64.0 & 87.0 & 51.5 & 80.0 & 58.3 & 88.8 & \textbf{54.1} & 83.4 \\
\bottomrule
\end{tabular}
}
\end{table*}

\begin{table*}[!htbp]
\centering
\caption{\textbf{Top-1 and Top-5 image retrieval accuracy (\%) across subjects for EEG encoder ablation studies}, including component ablations and pre-training transfer strategies, corresponding to Table~\ref{tab:ablation} and Table~\ref{tab:pretrain_ablation}, respectively. ``All Components'' is the final implementation of our EEG encoder and is included for comparison.}
\label{tab:eeg_ablation_study}
\scriptsize
\setlength{\tabcolsep}{3pt}
\renewcommand{\arraystretch}{1.2}
\resizebox{\textwidth}{!}{
\begin{tabular}{lcccccccccccccccccccccc}
\hline
\multirow{2}{*}{\textbf{Model}} 
& \multicolumn{2}{c}{Sub1} & \multicolumn{2}{c}{Sub2} & \multicolumn{2}{c}{Sub3} & \multicolumn{2}{c}{Sub4} & \multicolumn{2}{c}{Sub5}
& \multicolumn{2}{c}{Sub6} & \multicolumn{2}{c}{Sub7} & \multicolumn{2}{c}{Sub8} & \multicolumn{2}{c}{Sub9} & \multicolumn{2}{c}{Sub10} \\
\cmidrule(lr){2-3}\cmidrule(lr){4-5}\cmidrule(lr){6-7}\cmidrule(lr){8-9}\cmidrule(lr){10-11}\cmidrule(lr){12-13}\cmidrule(lr){14-15}\cmidrule(lr){16-17}\cmidrule(lr){18-19}\cmidrule(lr){20-21}
& Top-1 & Top-5 & Top-1 & Top-5 & Top-1 & Top-5 & Top-1 & Top-5 & Top-1 & Top-5
& Top-1 & Top-5 & Top-1 & Top-5 & Top-1 & Top-5 & Top-1 & Top-5 & Top-1 & Top-5 \\
\hline
\textbf{Spatial-Spectral}
& \textbf{57.5} & 84.3 & 52.0 & 80.7 & 44.8 & 78.7 & \textbf{59.2} & \textbf{90.5} & 42.7 & 75.7
& 50.2 & 82.2 & \textbf{50.5} & 80.0 & 67.7 & 88.3 & 46.3 & 77.0 & 59.5 & 90.0 \\

\textbf{Subject Layer}
& 54.5 & 85.0 & 50.8 & 79.2 & 47.0 & 78.0 & 55.0 & 89.2 & 45.5 & 78.8
& 49.5 & 80.3 & 48.3 & 79.7 & 67.2 & 87.3 & 45.2 & 76.7 & 58.2 & 88.3\\

\textbf{Transformer}
& 56.2 & 84.7 & 52.5 & 82.3 & 50.7 & \textbf{85.7} & 57.7 & 88.7 & \textbf{47.8} & \textbf{81.3}
& 50.7 & 83.3 & 41.0 & 77.8 & \textbf{67.7} & \textbf{88.7} & \textbf{54.5} & \textbf{81.3} & 59.3 & \textbf{90.5} \\

\textbf{GAT}
& 56.3 & 83.5 & 52.7 & 80.8 & 47.2 & 81.8 & 58.3 & 90.2 & 42.8 & 78.2
& \textbf{52.8} & \textbf{84.3} & 46.8 & 79.7 & 66.5 & 89.3 & 48.8 & 79.8 & 61.0 & 88.5\\

\hline
\textbf{None}

& 56.0 & 82.1 & 53.0 & 83.1 & 49.5 & 79.8 & 57.8 & 87.8 & 42.3 & 77.2 
& 47.1 & 78.2 & 46.1 & 78.3 & 64.8 & 86.7 & 49.7 & 76.7 & \textbf{60.2} & 89.6 \\

\textbf{All except Subject Layer}
& 56.7 & 82.3 & \textbf{53.3} & \textbf{83.7} & 49.2 & 80.3 & 59.0 & 88.5 & 42.7 & 77.3
& 48.2 & 77.7 & 47.2 & 79.2 & 63.7 & 86.5 & 48.7 & 76.0 & 58.7 & 89.0 \\

All Components
& 56.5 & \textbf{85.5} & 52.3 & 81.8 & \textbf{53.3} & 79.7 & 56.7 & 86.7 & \textbf{47.5} & 80.5 & 50.3 & 83.3 & \textbf{50.0} & \textbf{80.3} & 64.0 & 87.0 & 51.5 & \textbf{80.0} & 58.3 & 88.8  \\

\hline

\end{tabular}%
}
\end{table*}

\begin{table*}[!htbp]
\centering
\caption{\textbf{Prompt format and partial examples of LLM-generated visual descriptions} (generated using the Qwen2-VL-7B model).}
\label{tab:llm_prompt_examples}
\renewcommand{\arraystretch}{1.2}
\small
\begin{tabular}{p{0.95\linewidth}}
\hline
\textbf{Prompt} \\
\hline
\ttfamily
Role: user \\
Content: \\
\ \ - type: image, image: image \\
\ \ - type: text, text: Describe only what is directly visible in the image of \textless label\textgreater\ in one short sentence. 
\normalfont
\\
\hline
\textbf{Partial LLM-generated results} \\
Label: aardvark \\
Content: The aardvark has light-brown fur, large ears, and dark legs with black feet. It appears to be walking on grassy ground. \\[4pt]
Label: airbag \\
Content: The image shows an inflated airbag inside a car, covering part of the steering wheel and dashboard. \\[4pt]
Label: airboat \\
Content: A blue airboat with a canopy and propellers is positioned on grass, facing left against a clear sky backdrop. \\[4pt]
Label: abacus \\
Content: The image shows an abacus with colorful beads, including red, green, yellow, and white ones, arranged on parallel wires within a wooden frame. \\
\hline
\end{tabular}
\end{table*}

\FloatBarrier
\clearpage
\section{Supplementary Visualizations}
\label{app:supplementary_visualizations}

\begin{figure}[!htbp]
    \centering
    \includegraphics[width=0.7\linewidth]{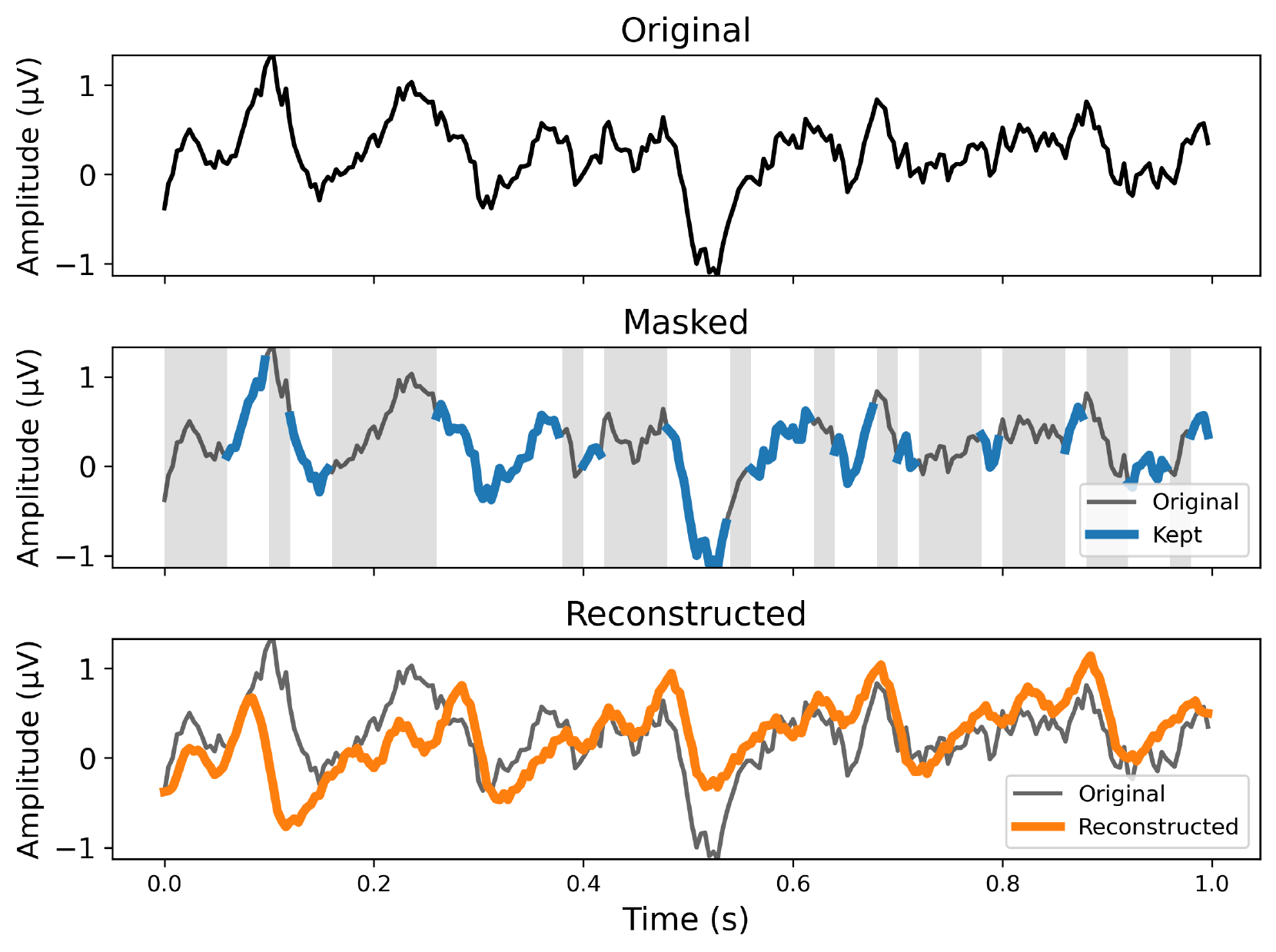}
    \caption{An example of \textbf{masked EEG input and reconstructed result} for a randomly selected channel from one trial. The reconstructed waveform captures the main low-frequency trends of the original signal, while fine-grained details remain limited by the inherent noise of EEG recordings.}
    \label{fig:mae}
\end{figure}

\begin{figure}[!htbp]
    \centering
    \includegraphics[width=0.65\linewidth]{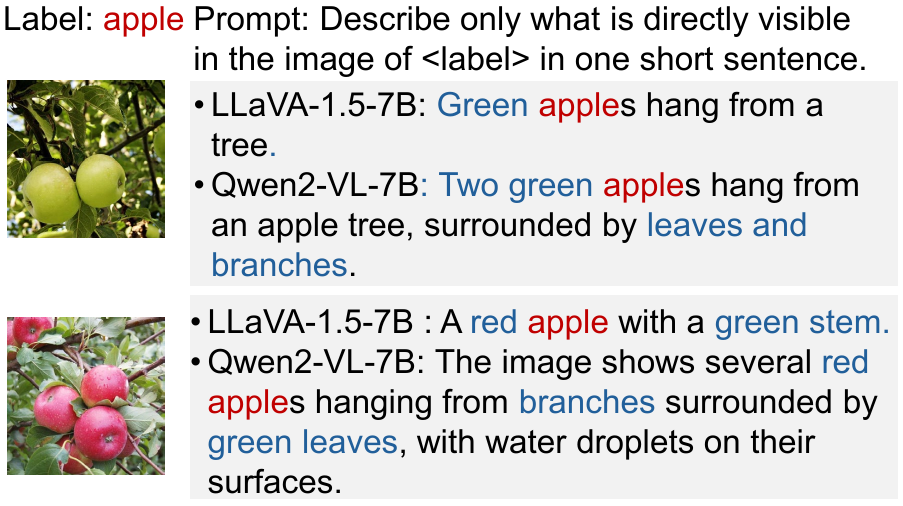}
    \caption{\textbf{Descriptions generated for an image using different LLMs}. Red indicates the object label, and blue indicates object details. Qwen2-VL-7B generates more detailed, context-rich descriptions, capturing attributes such as quantity and surrounding elements, whereas LLaVA-1.5-7B tends to produce more concise descriptions focused on the primary object.
}
    \label{fig:llm and prompt}
\end{figure}

\begin{figure}[!htbp]
    \centering
    \includegraphics[width=0.9\linewidth]{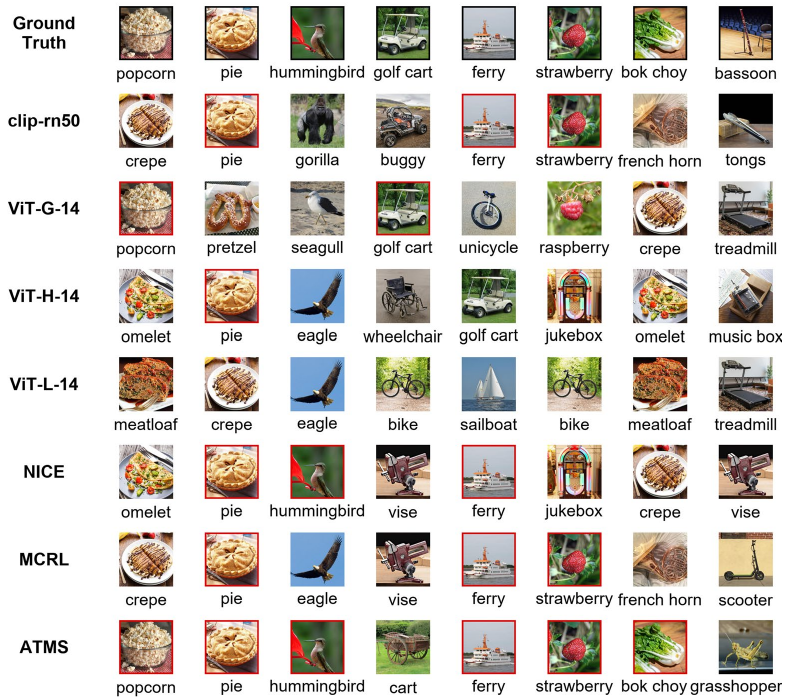}
    \caption{\textbf{Qualitative Top-1 retrieval results obtained with different visual and EEG encoders}, with the ground-truth image shown in the first row. The results are generated following the same experimental configuration as those evaluated in Tables~\ref{tab:image_encoder_subjects} and \ref{tab:top1_top5_by_encoder}}
    \label{fig:retrieval_examples}
\end{figure}

\begin{figure}[!htbp]
    \centering
    \includegraphics[width=1\linewidth]{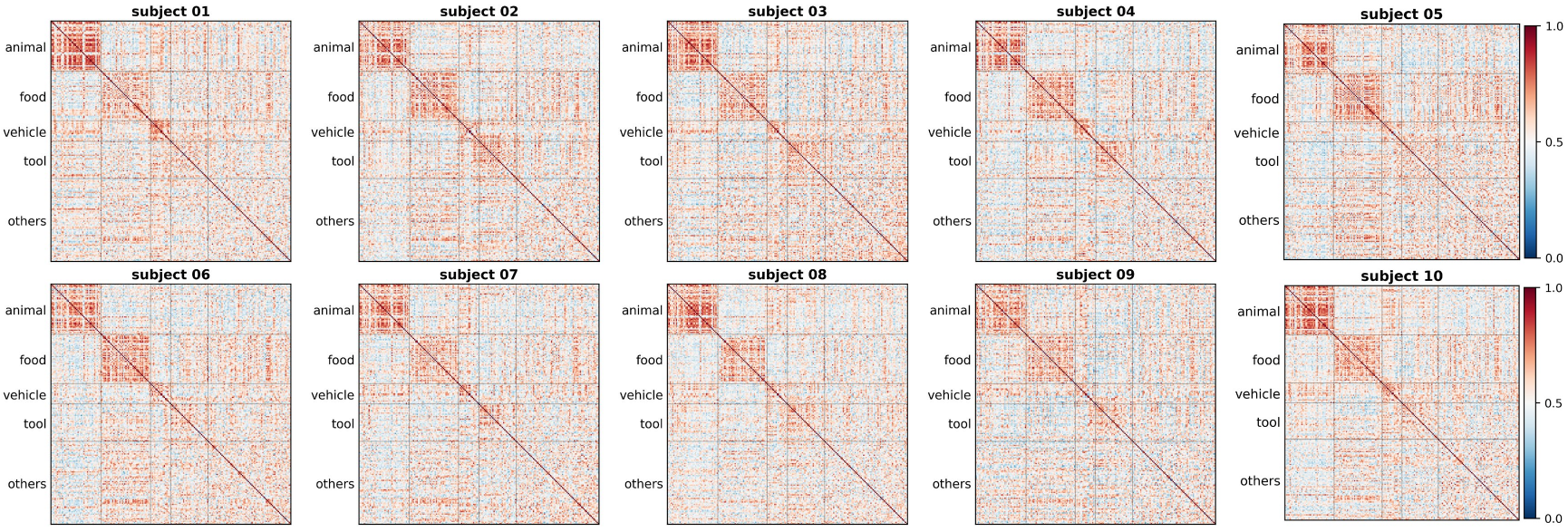}
    \caption{\textbf{Representational similarity matrices} across 10 subjects.}
    \label{fig:10subsimilarity}
\end{figure}

\begin{figure}[!htbp]
    \centering
    \includegraphics[width=1\linewidth]{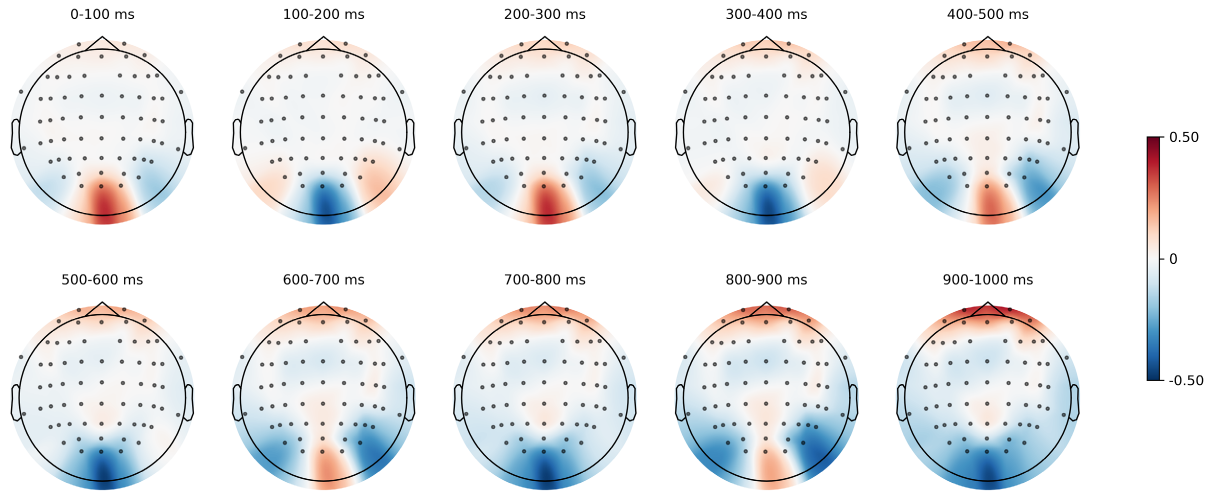}
    \caption{\textbf{Topographies of EEG signals averaged across all trials for Subject 1 at 100 ms intervals}. A clear response is observed in the occipital area (0-100 ms), followed by activity in the temporal area (100-600 ms) after stimulus onset. The 200-ms SOA still induces periodic responses in the occipital cortex. Frontal activity gradually increases, possibly reflecting additional cognitive processes.}
    \label{fig:topographies}
\end{figure}

We provide additional qualitative visualizations: a masked 
EEG input/reconstruction example (Fig.~\ref{fig:mae}), LLM-generated 
descriptions for sample images (Fig.~\ref{fig:llm and prompt}), 
qualitative Top-1 retrievals across encoders (Fig.~\ref{fig:retrieval_examples}), 
per-subject representational similarity matrices 
(Fig.~\ref{fig:10subsimilarity}), and EEG topographies for Subject~1 
(Fig.~\ref{fig:topographies}).

\FloatBarrier

 
\end{document}